\documentclass[preprint,review,11pt]{elsarticle}

\usepackage{amssymb}
\usepackage{amsmath,amsfonts}
\usepackage{algorithmic}
\usepackage{algorithm}
\usepackage{array}
\usepackage{caption}
\usepackage{subcaption}
\usepackage{textcomp}
\usepackage{stfloats}
\usepackage{url}
\usepackage{verbatim}
\usepackage{graphicx}
\usepackage{array}
\newcolumntype{P}[1]{>{\centering\arraybackslash}p{#1}}
\usepackage{comment}
\usepackage{natbib}
\setcitestyle{square}

\biboptions{sort&compress}
\usepackage{tabularx,ragged2e,booktabs,caption}
\usepackage{multirow}
\usepackage[export]{adjustbox}
\usepackage [latin1]{inputenc}
\usepackage{acronym}
\usepackage{setspace}

\usepackage{lineno}

\usepackage[dvipsnames]{xcolor}
\newif\ifFinalVer
\FinalVertrue
\newcommand{\colorOpt}[2]{\ifFinalVer{#2}\else{{\color{#1}#2}}\fi}

\newcommand{\figureTwoLevelFlowChart}{
	\begin{figure*}[htbp]
		\centering
		\includegraphics[width = \textwidth]{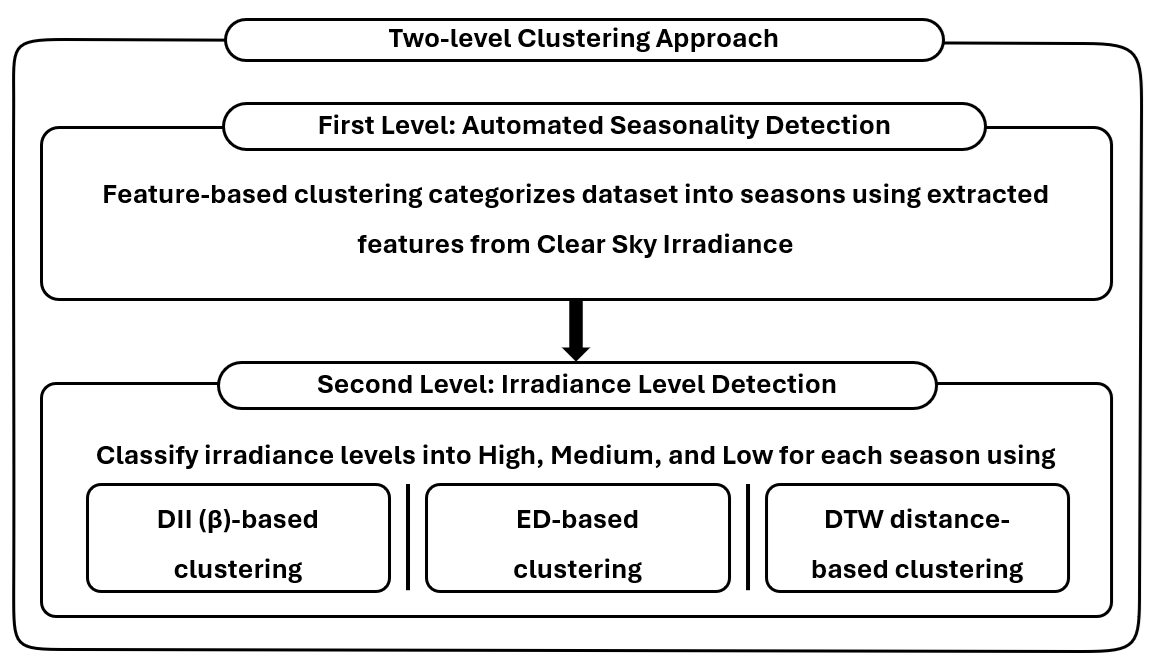}
		\caption{Two-level clustering framework for irradiance analysis.}
		\label{fig:twoLevelFlowChart}
	\end{figure*}
}

\newcommand{\figureCloudCoverSamples}{
	\begin{figure}
		\centering
		\begin{subfigure}{\columnwidth}
			\centering
			\includegraphics[width=0.9\textwidth]{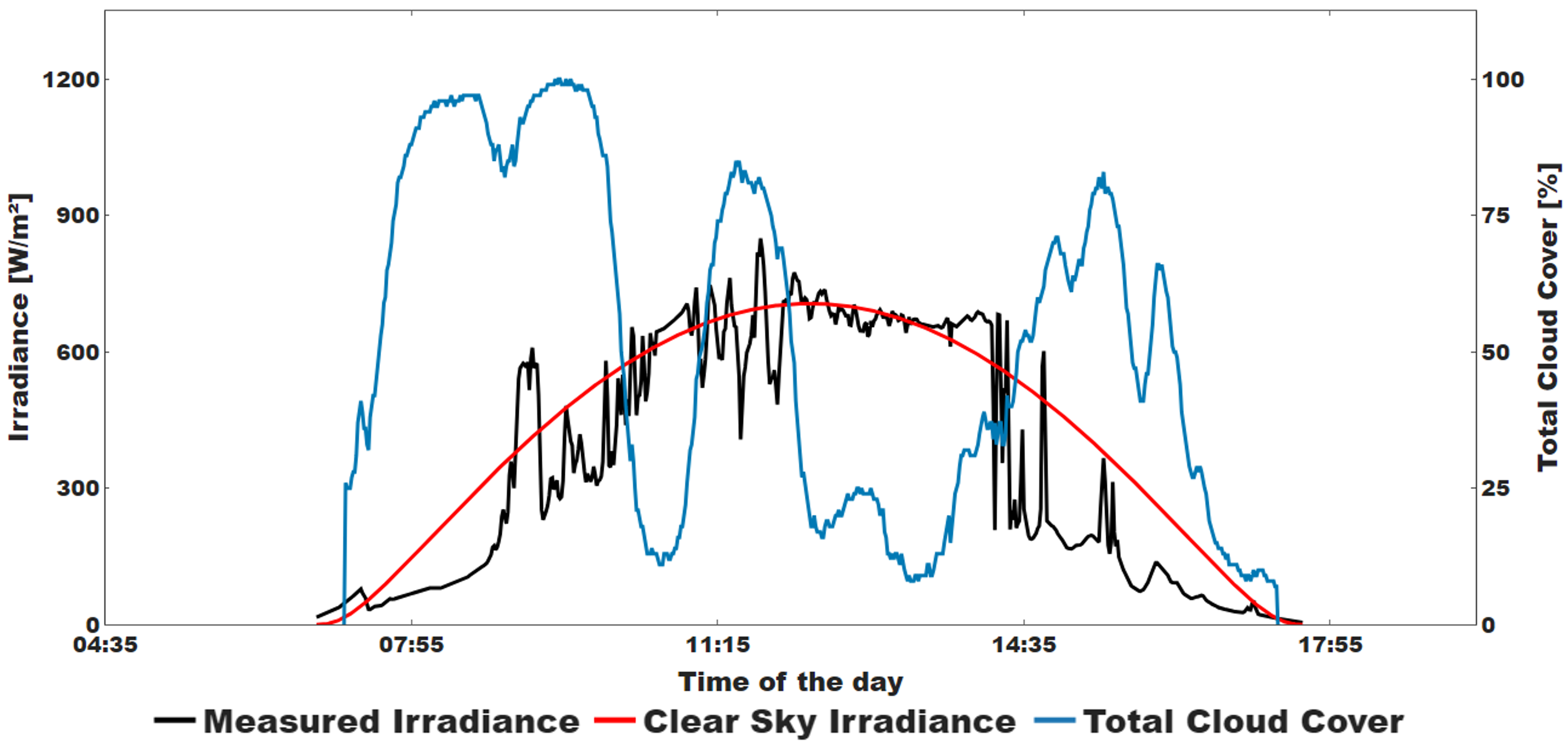}
			\caption{}
		\end{subfigure}
		
		\medskip
		
		\begin{subfigure}{0.45\textwidth}
			\includegraphics[width=\textwidth]{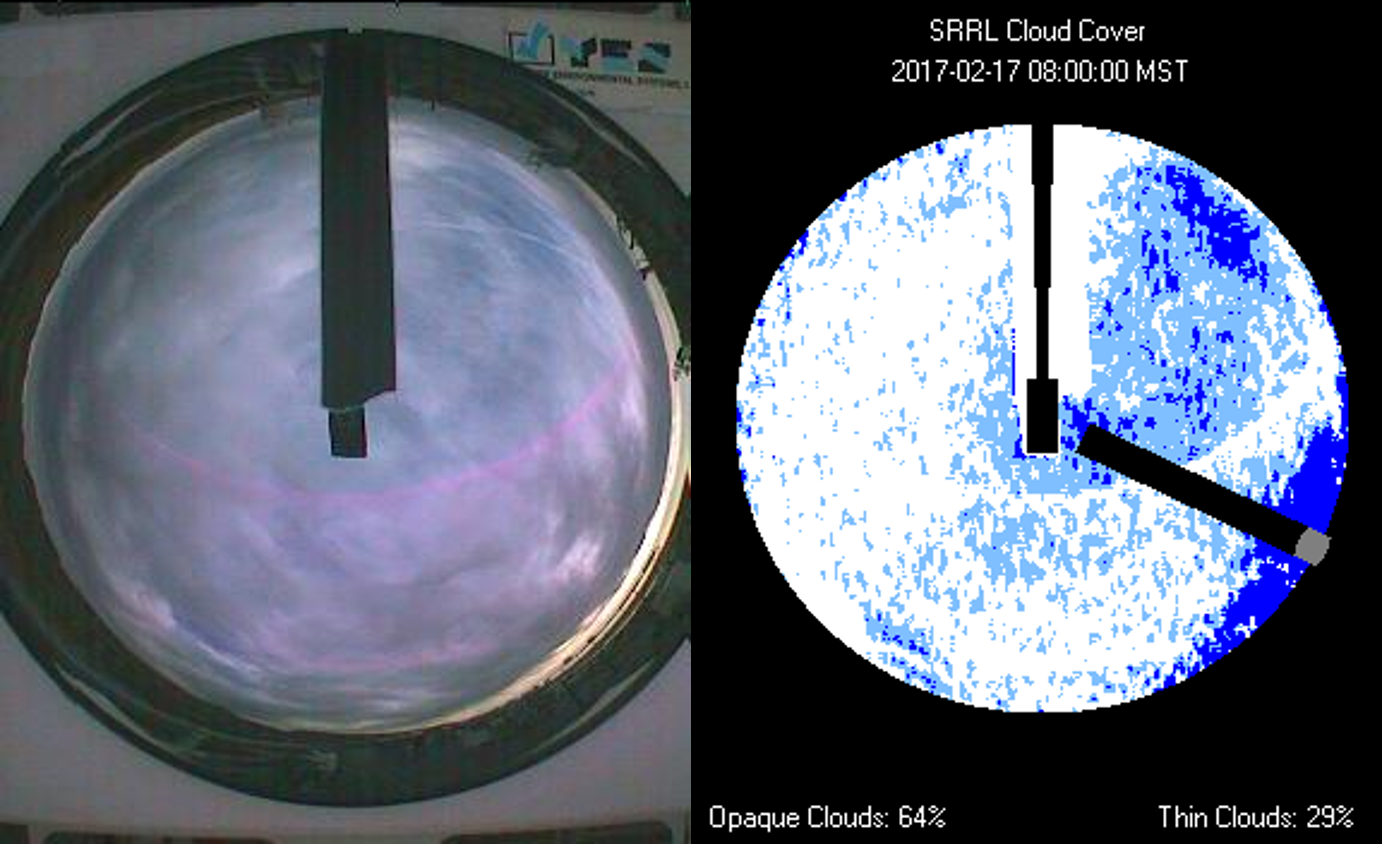}
			\caption{At 8:00, Total CC = 93\%, Opaque CC = 64\%, and Thin CC = 29\%}
		\end{subfigure}
		\hfill
		\begin{subfigure}{0.45\textwidth}
			\includegraphics[width=\textwidth]{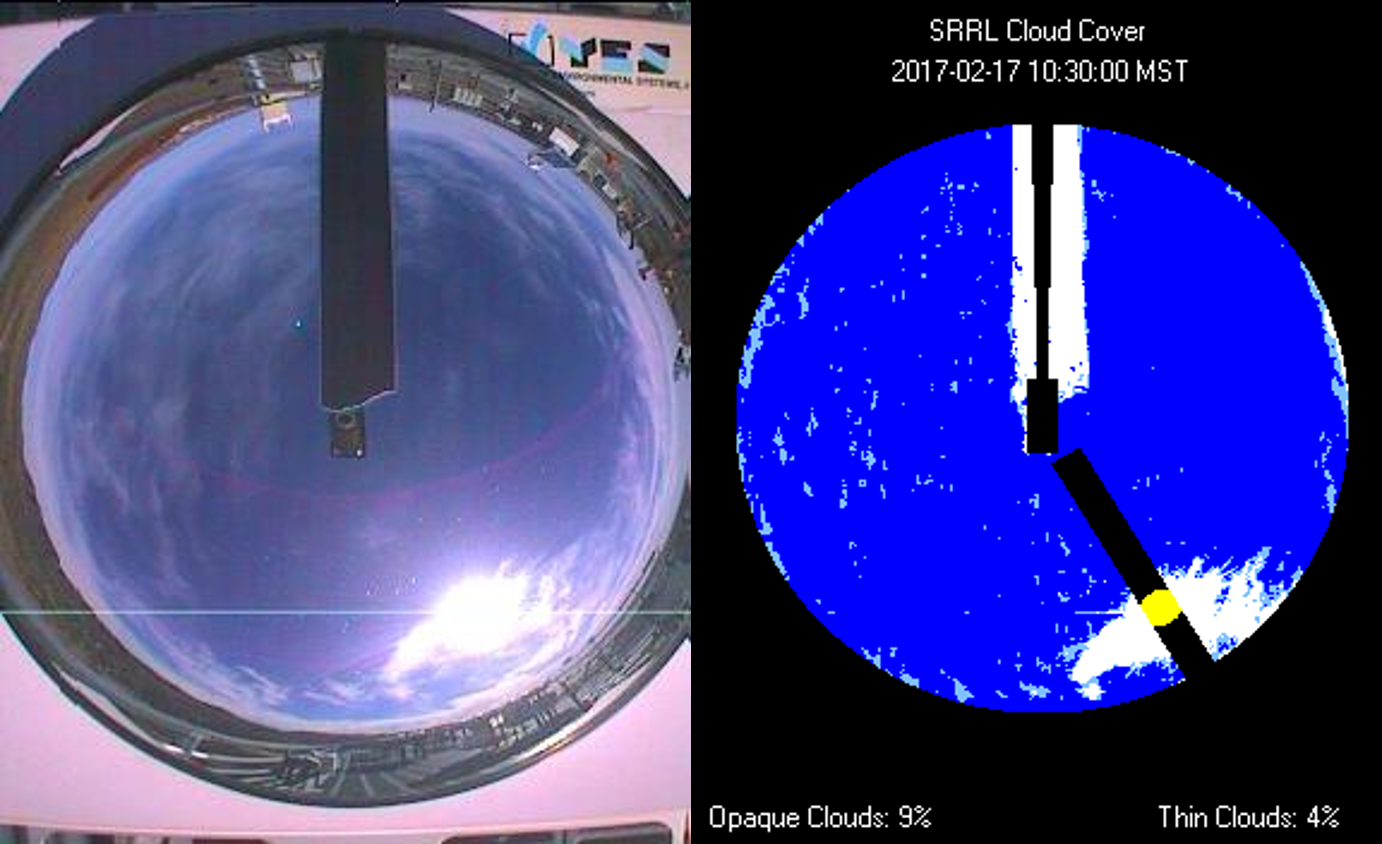}
			\caption{At 10:30, Total CC = 13\%, Opaque CC = 9\%, and Thin CC = 4\%}
		\end{subfigure}
		
		\medskip
		
		\begin{subfigure}{0.45\textwidth}
			\includegraphics[width=\textwidth]{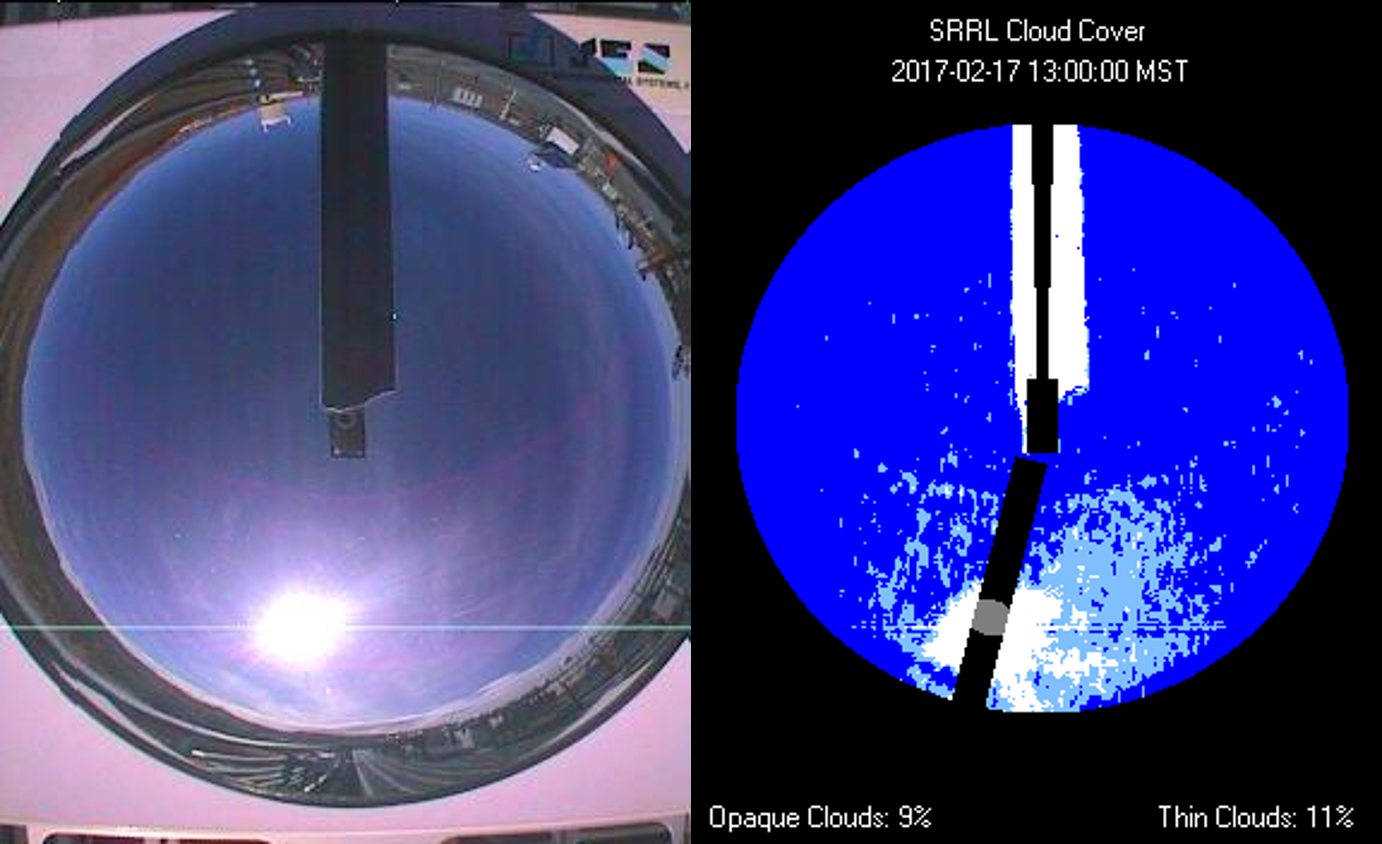}
			\caption{At 13:00, Total CC = 20\%, Opaque CC = 9\%, and Thin CC = 11\%}
		\end{subfigure}
		\hfill
		\begin{subfigure}{0.45\textwidth}
			\includegraphics[width=\textwidth]{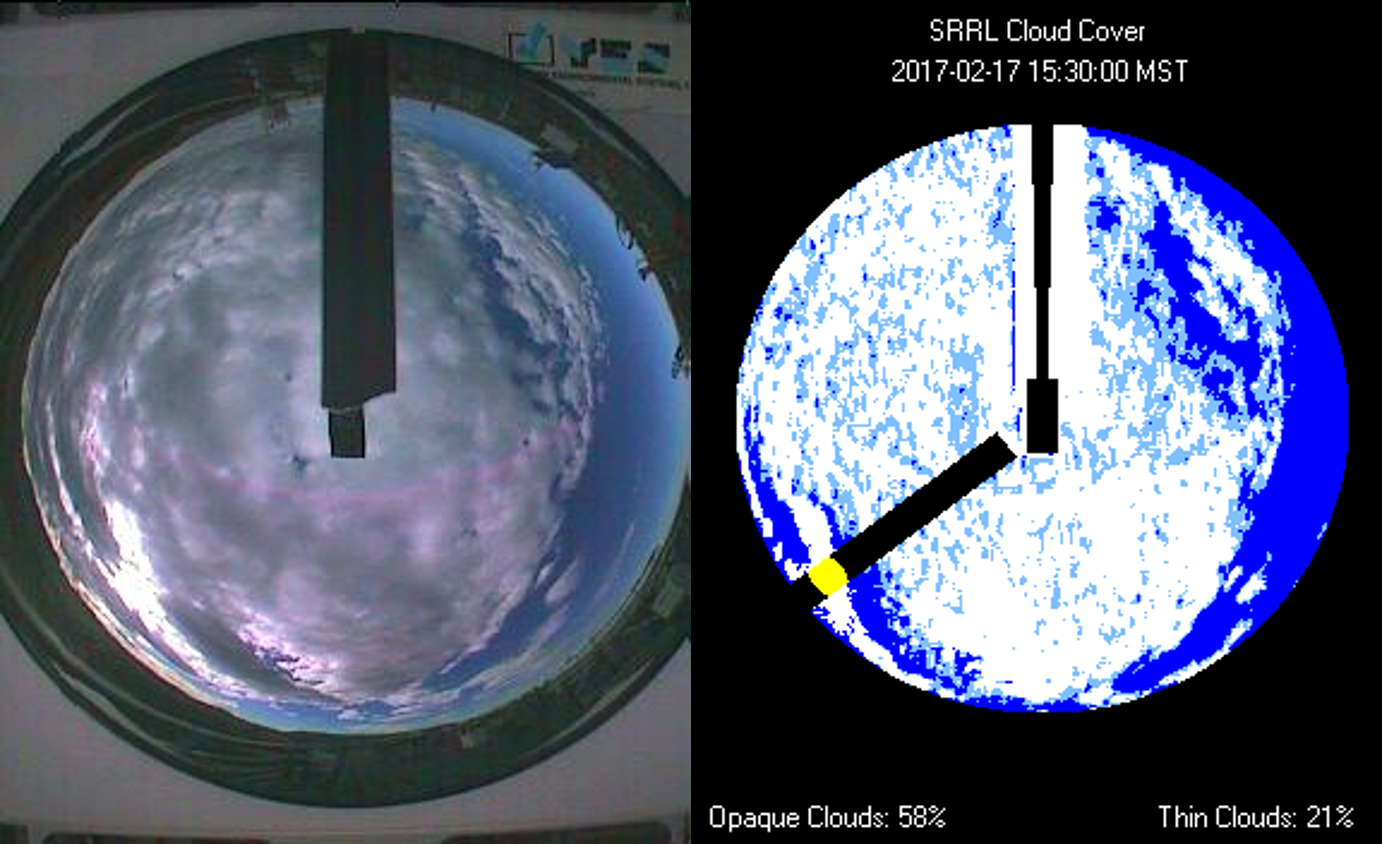}
			\caption{At 15:30, Total CC = 79\%, Opaque CC = 58\%, and Thin CC = 21\%}
		\end{subfigure}
		
		\caption{(a) Comparison of measured irradiance profiles of 17 February 2017 with  the  measured irradiance, clear  sky  irradiance  superimposed  with total cloud cover measurements (right Y-axis), (b) - (e) All-sky cloud cover images at given sampling time with raw image on left and corresponding cloud decision image on the right with thin cloud cover [\%] and opaque cloud cover [\%]}
		\label{fig:cloudcoversamples}
		
	\end{figure}
}

\newcommand{\figureSeasonsHawaiiSubplots}{
	\begin{figure*}[tbh]
		\begin{subfigure}{0.2\textwidth}
			\centering
			\includegraphics[width=\textwidth]{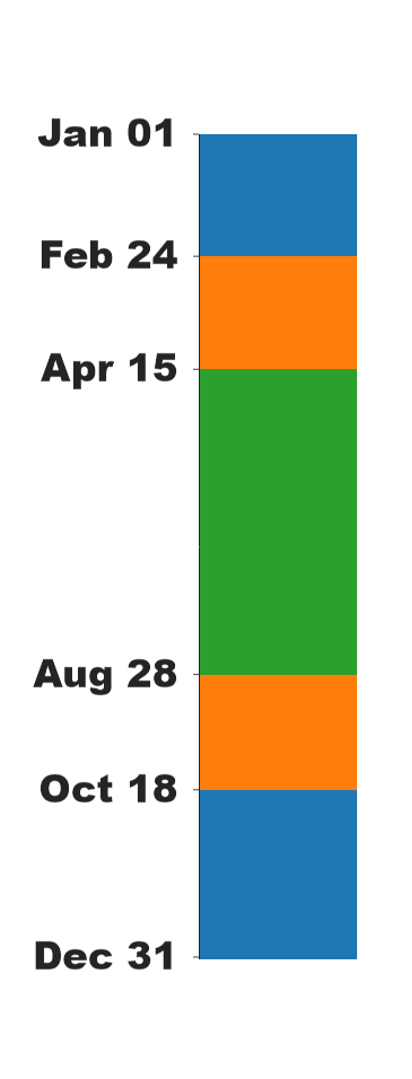}
			\caption{}
		\end{subfigure}
		\begin{subfigure}{0.78\textwidth}
			\centering
			\includegraphics[width=\textwidth]{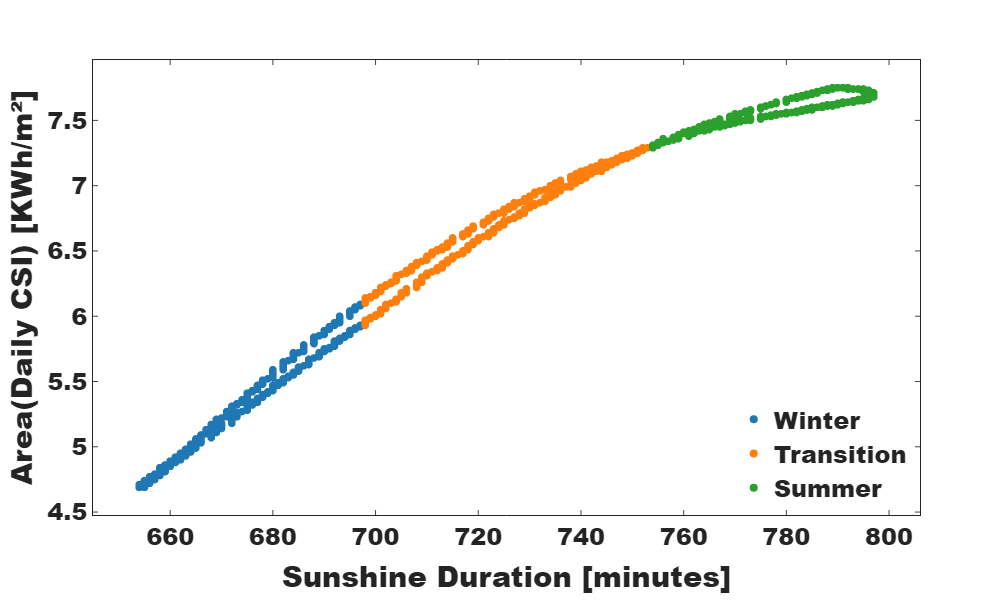}
			\caption{}
		\end{subfigure}
		\caption{From the first level clustering for Hawaii, (a) identified seasonal boundaries, and (b) identified seasonal clusters.}
		\label{fig: season scatter plot Hawaii and seasonal boundary with vertical colorbar}
	\end{figure*}
}

\newcommand{\figureBetaCloudCover}{
	\begin{figure*}[htbp]
		\centering
		\includegraphics[width = \textwidth]{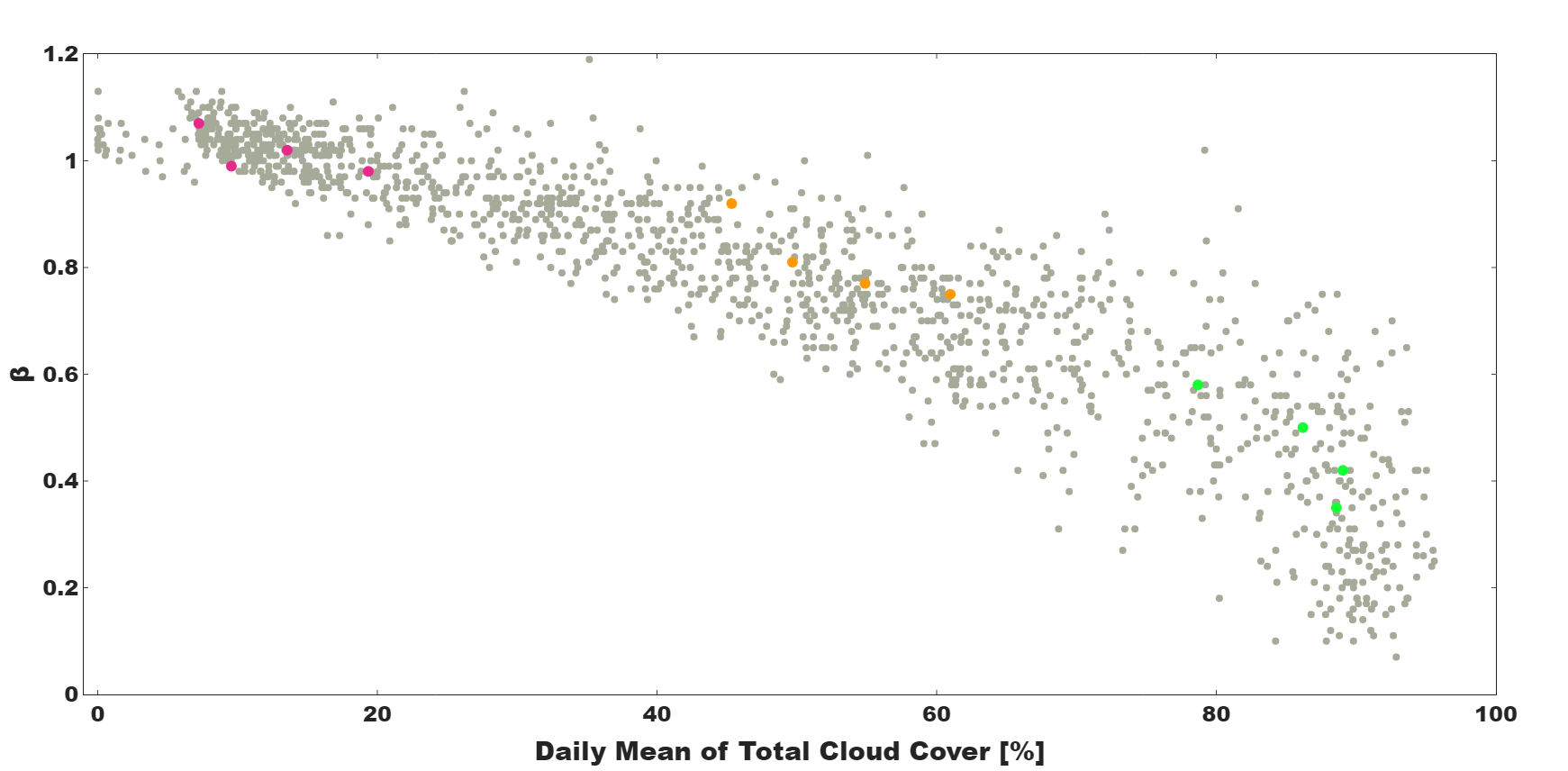}
		\caption{Scatter plot of $\beta$ and daily mean of total cloud cover obtained from the measurements. Profiles of selected days, shown as pink, orange, and green-filled circles, are plotted in Figure \ref{fig:betasamplesHighBeta}, \ref{fig:betasamplesMedBeta}, and \ref{fig:betasamplesLowBeta}, respectively.}
		\label{fig:betasdistribution}
	\end{figure*}
}

\newcommand{\figureSampledIrrAndCloudCover}{
\begin{figure}[htbp]
     \centering
     \begin{subfigure}[b]{\columnwidth}
         \centering
         \includegraphics[width = 0.8\textwidth]{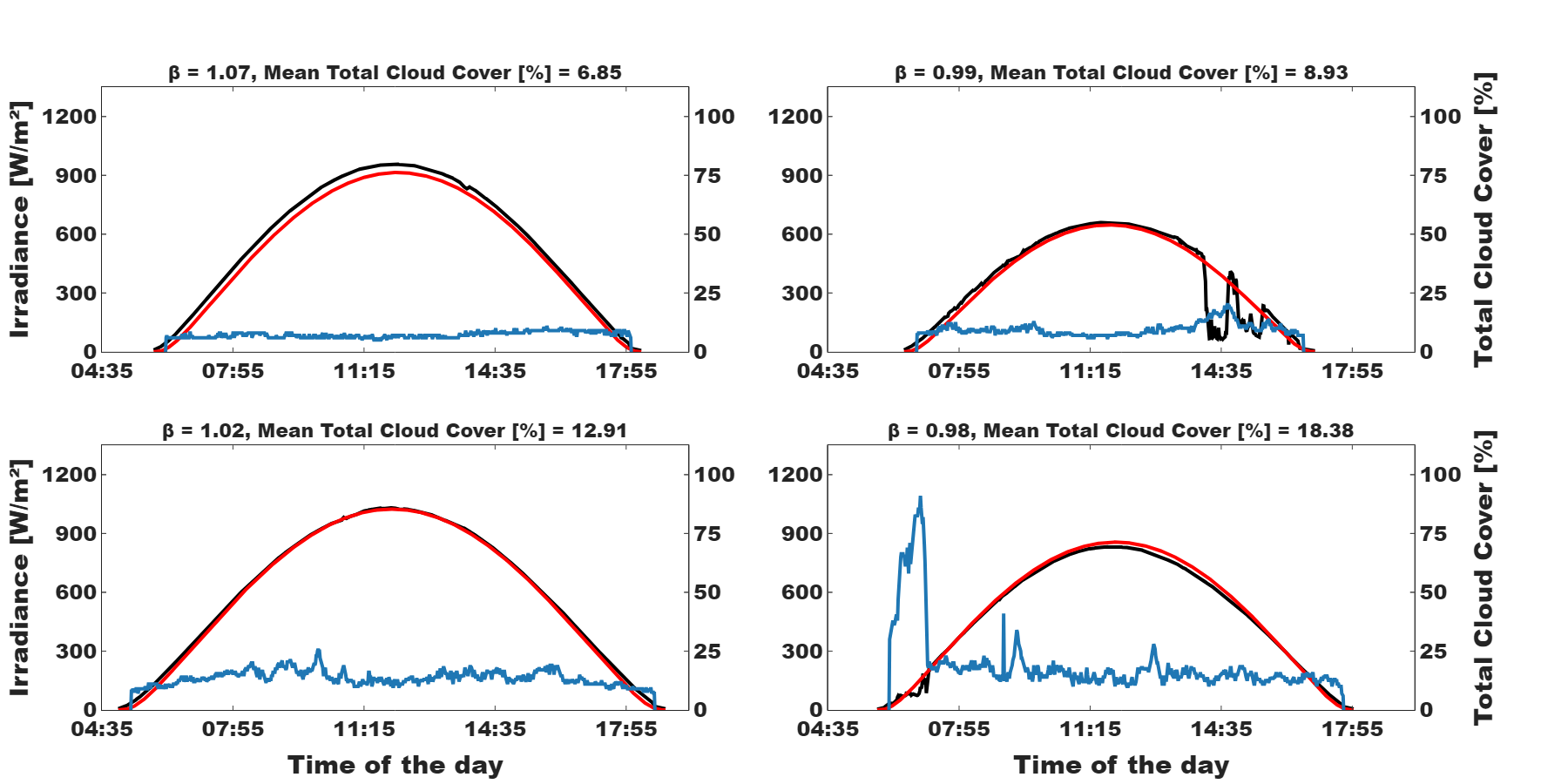}
         \caption{}
         \label{fig:betasamplesHighBeta}         
     \end{subfigure}
     \centering
     \begin{subfigure}[b]{\columnwidth}
         \centering
         \includegraphics[width = 0.8\textwidth]{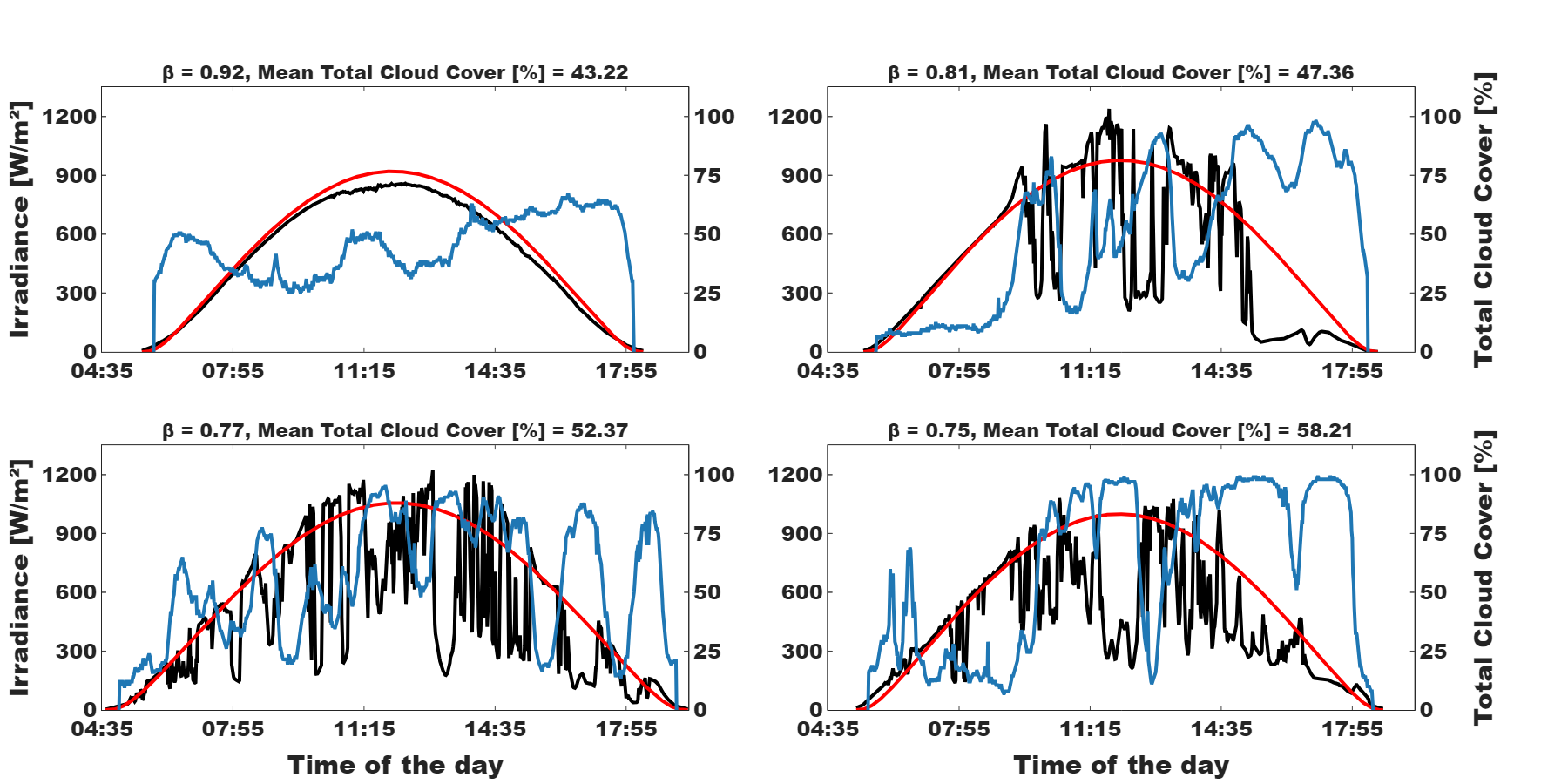}
         \caption{}
         \label{fig:betasamplesMedBeta}         
     \end{subfigure}
     \centering
     \begin{subfigure}[b]{\columnwidth}
         \centering
         \includegraphics[width = 0.8\textwidth]{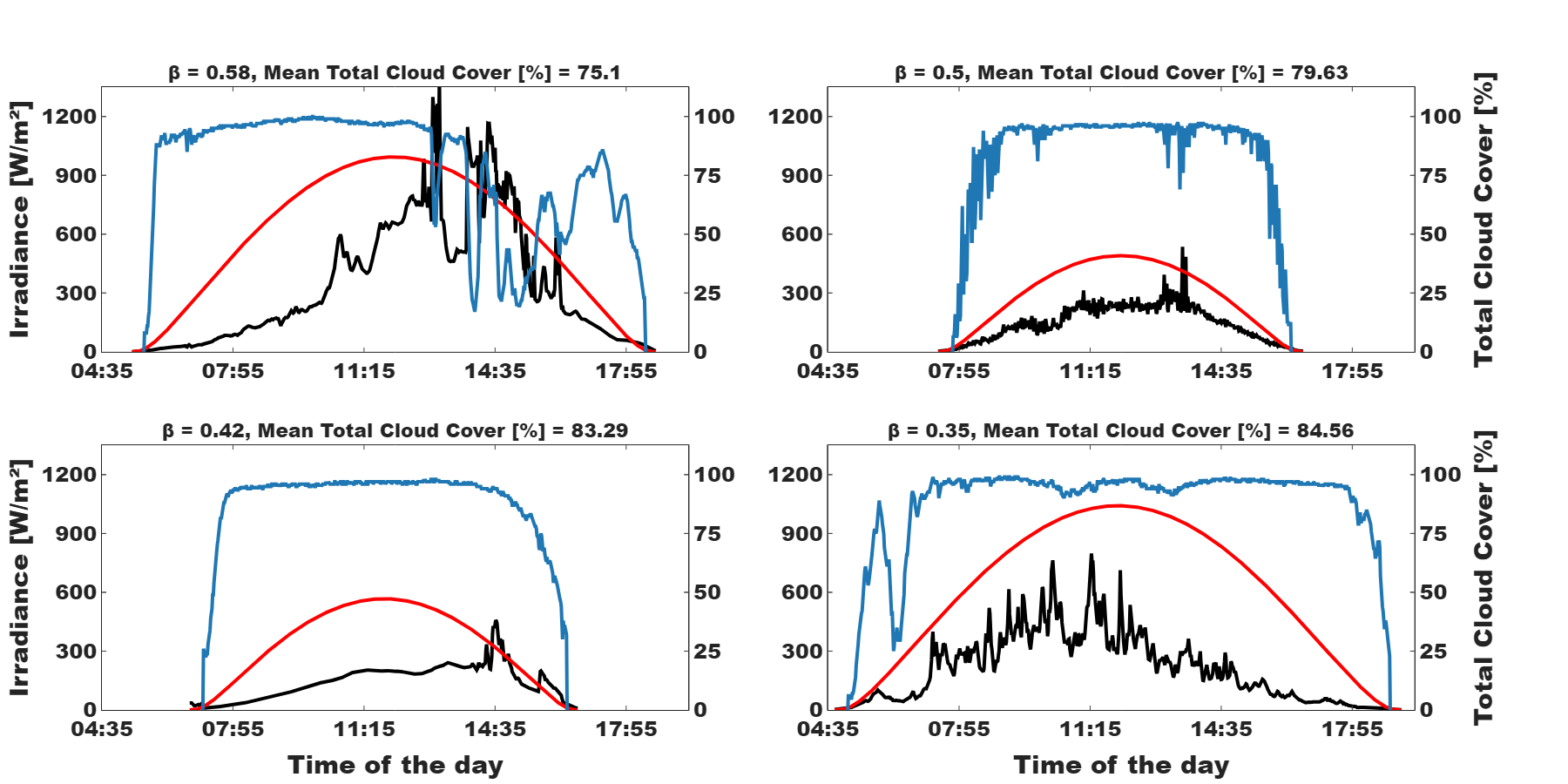}
         \caption{}
         \label{fig:betasamplesLowBeta}         
     \end{subfigure}
     \centering
     \begin{subfigure}[b]{\columnwidth}
         \centering
         \includegraphics[width = 0.6\textwidth]{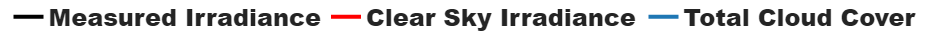}
     \end{subfigure}
\caption{Comparison of measured irradiance profiles of selected days with the clear sky irradiance superimposed with cloud cover measurements (right Y-axis): (a) low cloud cover; high $\beta$, (b) medium cloud cover; medium $\beta$, and (c) high cloud cover; low $\beta$. The selected days are represented by pink, orange, and green-filled circles in Figure \ref{fig:betasdistribution}.}
\label{fig:betasamples}
\end{figure}
}

\newcommand{\figureBetaMonthlyYearOnYearGoldenHawaii}{
	\begin{figure}[htbp]
		\centering
		\begin{subfigure}[b]{\columnwidth}
			\centering
			\includegraphics[width = \columnwidth]{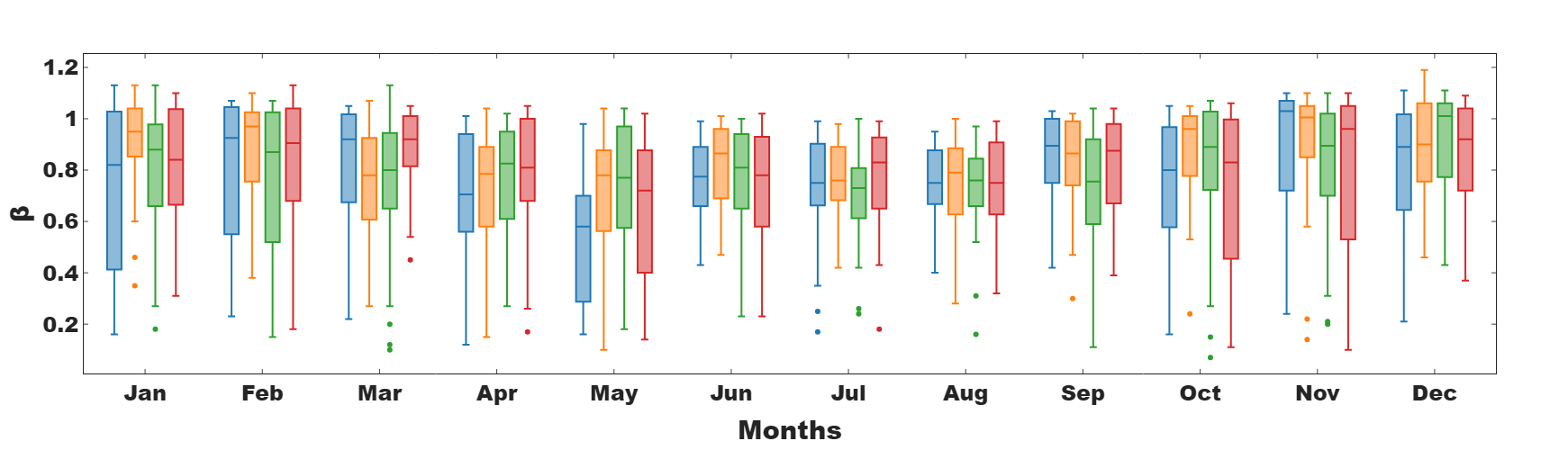}
			\caption{}
		\end{subfigure}
		\begin{subfigure}[b]{\columnwidth}
			\centering
			\includegraphics[width = \columnwidth]{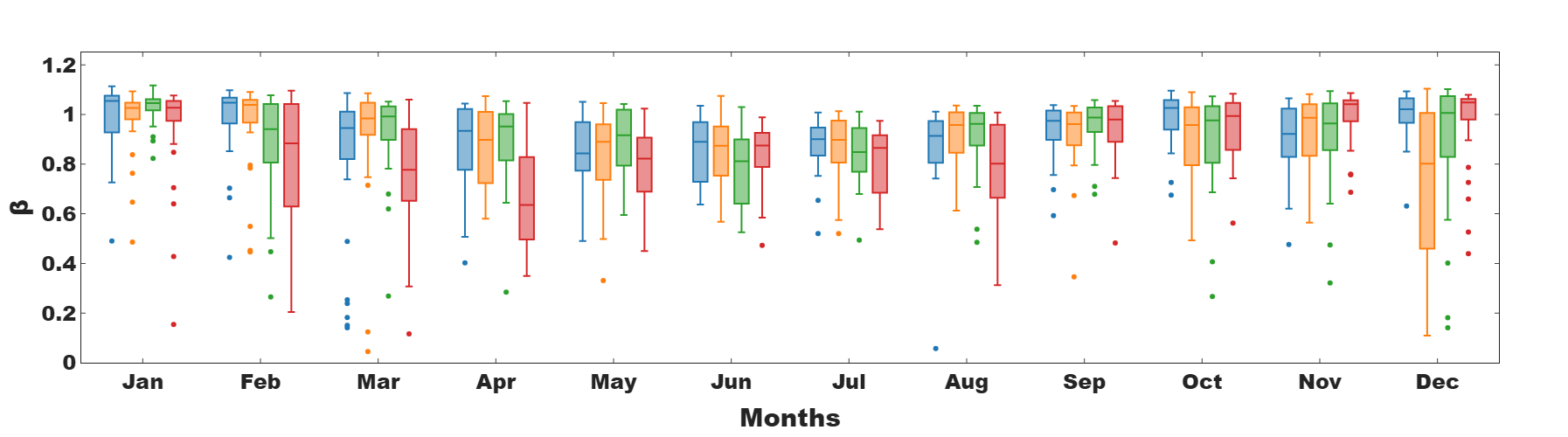}
			\caption{}
			\label{fig:betaMonthlyGoldenHawaii}
		\end{subfigure}
		\caption{Box-Whisker plot showing the $\beta$ distribution for all the calendar months for over the four years analyzed for (a) Golden and (b) Hawaii. The blue, yellow, green, and red box plots correspond to the years 2015, 2016, 2017, and 2018, respectively.}
		\label{fig:betaMonthlyGoldenHawaii}
	\end{figure}
}

\newcommand{\figureBetaEDDTWMeanProfilesGolden}{
	\begin{figure}[htbp]
		\centering
		\begin{subfigure}[b]{\columnwidth}
			\centering
			\includegraphics[width = 0.8\textwidth]{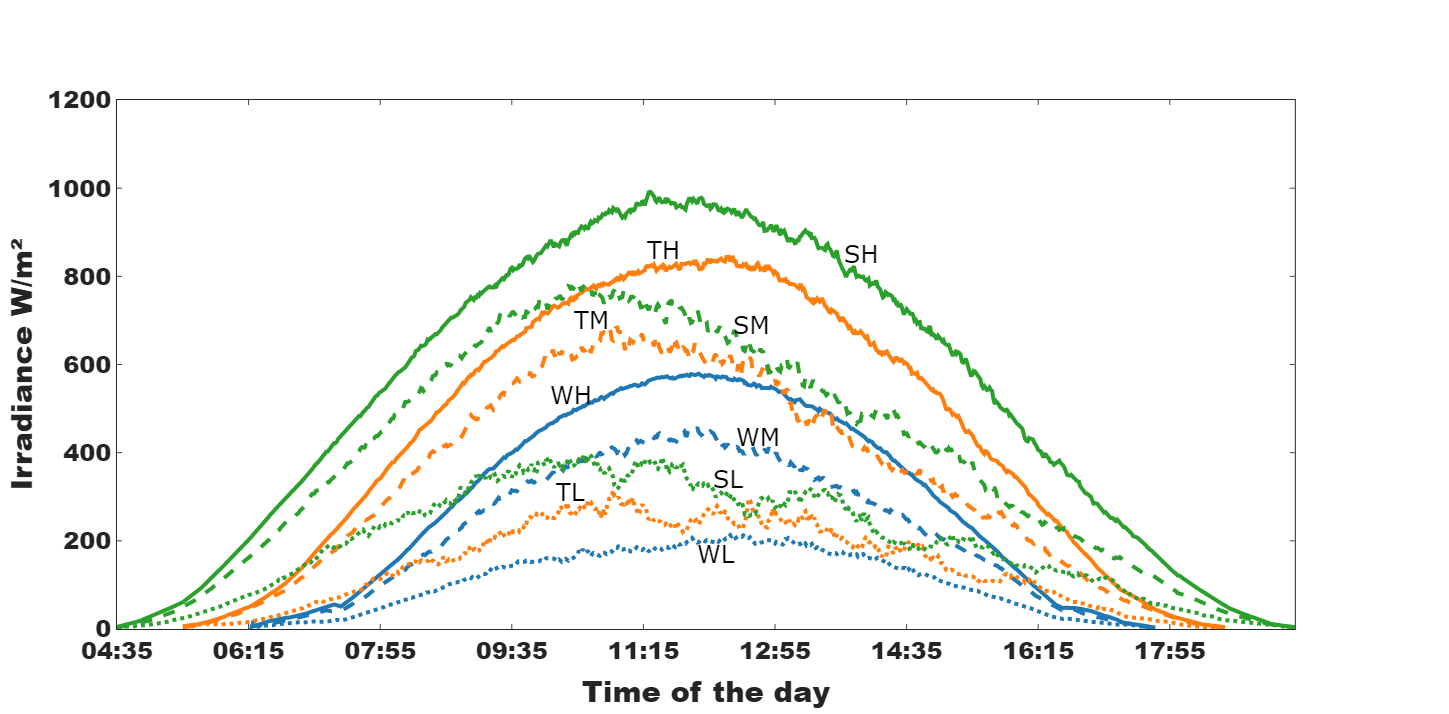}
			\caption{}
			\label{fig: beta - mean profiles}
		\end{subfigure}
		\begin{subfigure}[b]{\columnwidth}
			\centering
			\includegraphics[width = 0.8\textwidth]{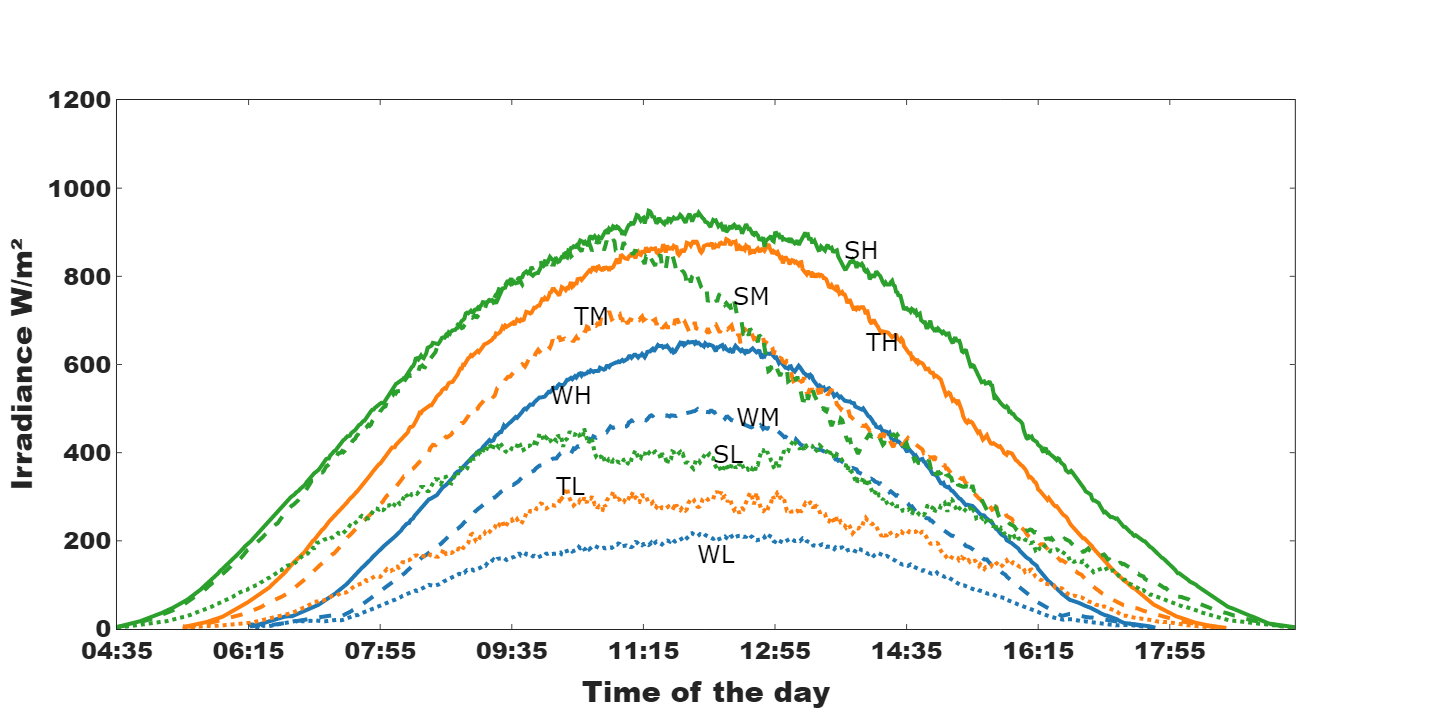}
			\caption{}
			\label{fig: ED mean profiles}
		\end{subfigure}
		\begin{subfigure}[b]{\columnwidth}
			\centering
			\includegraphics[width = 0.8\textwidth]{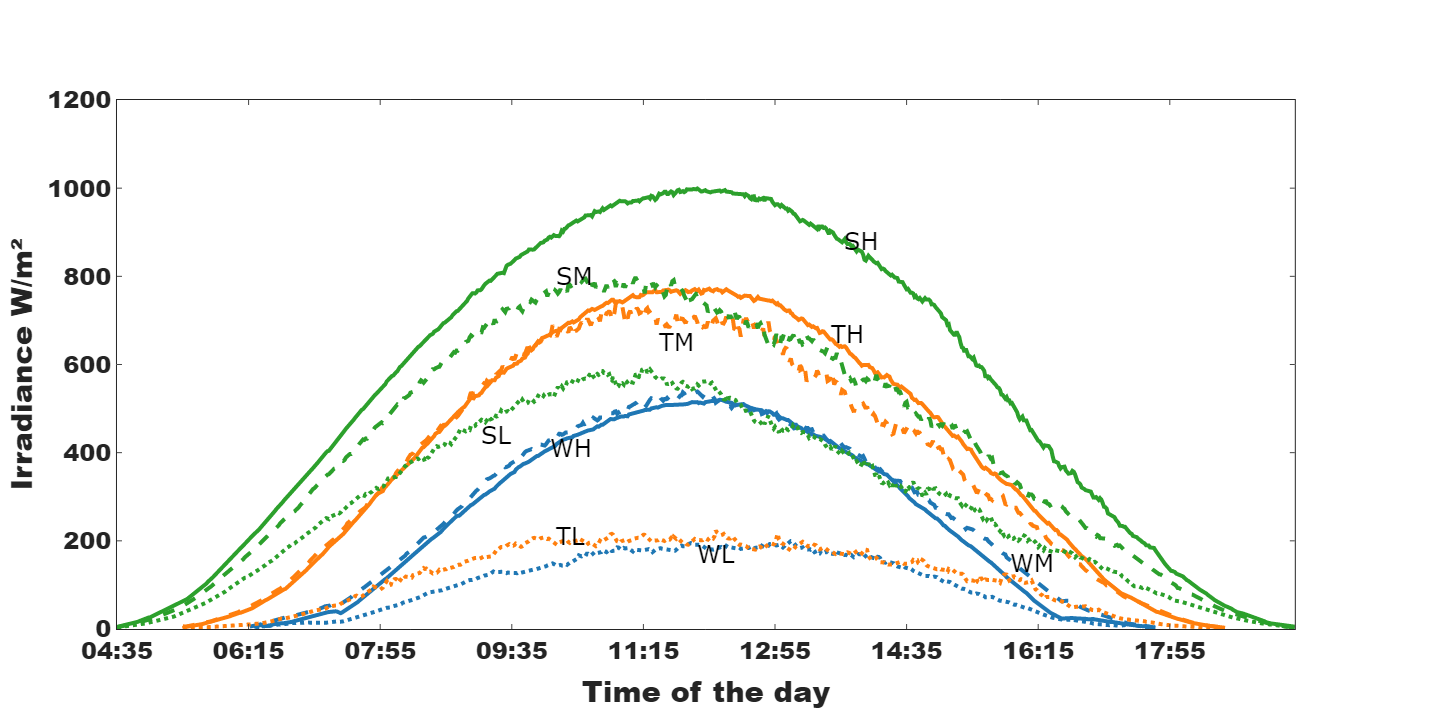}
			\caption{}
			\label{fig:DTW mean profiles}
		\end{subfigure}
		\caption{Cluster mean irradiance profiles obtained by different clustering approaches for Golden: (a) $\beta$-based, (b) ED time-series, and (c) DTW time-series. The first letter in the label represents the seasonal cluster (S, T or W), and the second letter represents the irradiance level cluster (H, M and L). The seasons S, T and W are shown in green, orange and blue lines, while the H, M, and L clusters in each season as solid, dashed, and dotted lines corresponding to clear, partially cloudy, and cloudy days, respectively.}
		\label{fig: mean profiles Golden}
	\end{figure}
}

\newcommand{\figureBetaEDDTWMeanProfilesHawaii}{
	\begin{figure}[htbp]
		\centering
		\begin{subfigure}[b]{\columnwidth}
			\centering
			\includegraphics[width = 0.9\textwidth]{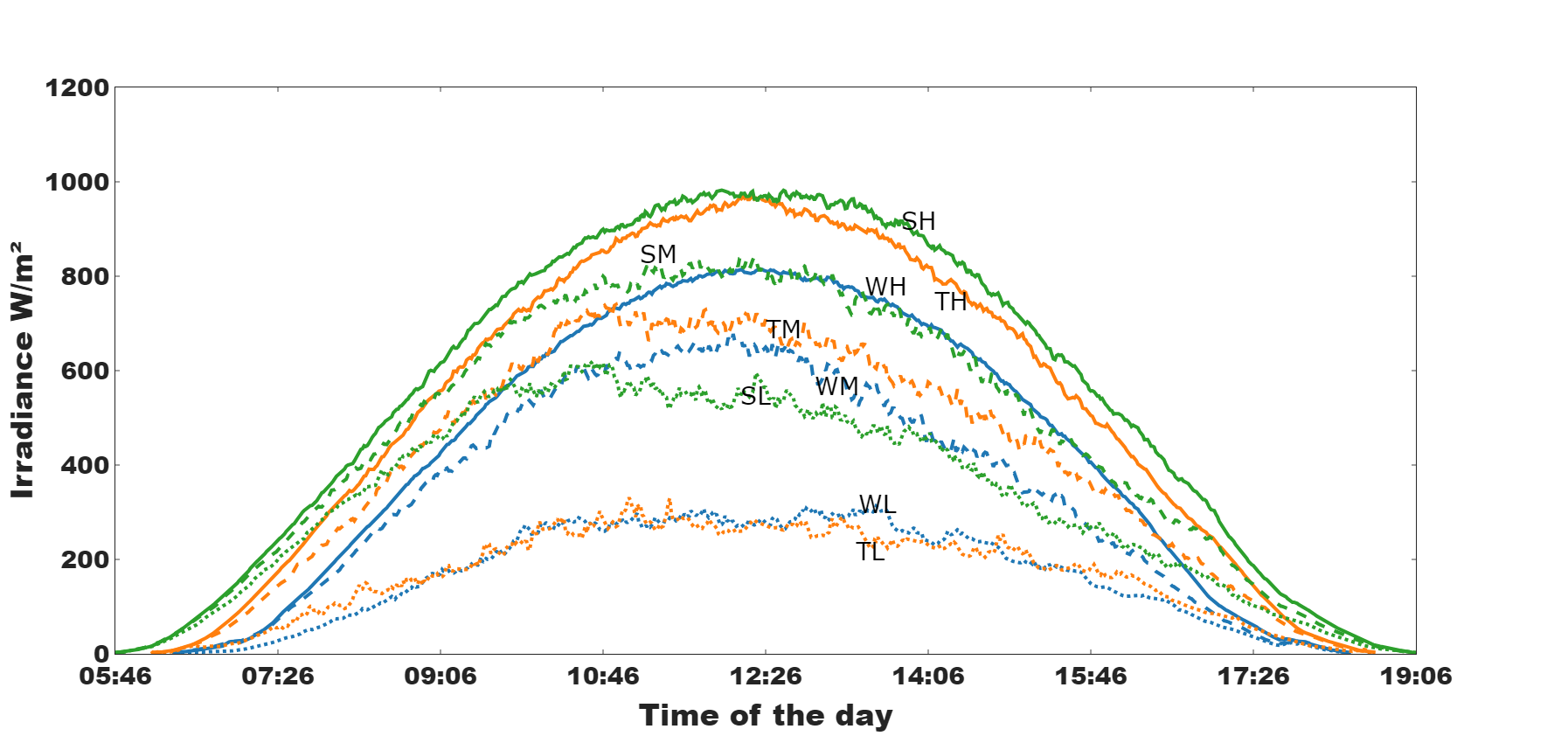}
			\caption{}
			\label{fig: beta - mean profiles}
		\end{subfigure}
		\begin{subfigure}[b]{\columnwidth}
			\centering
			\includegraphics[width = 0.9\textwidth]{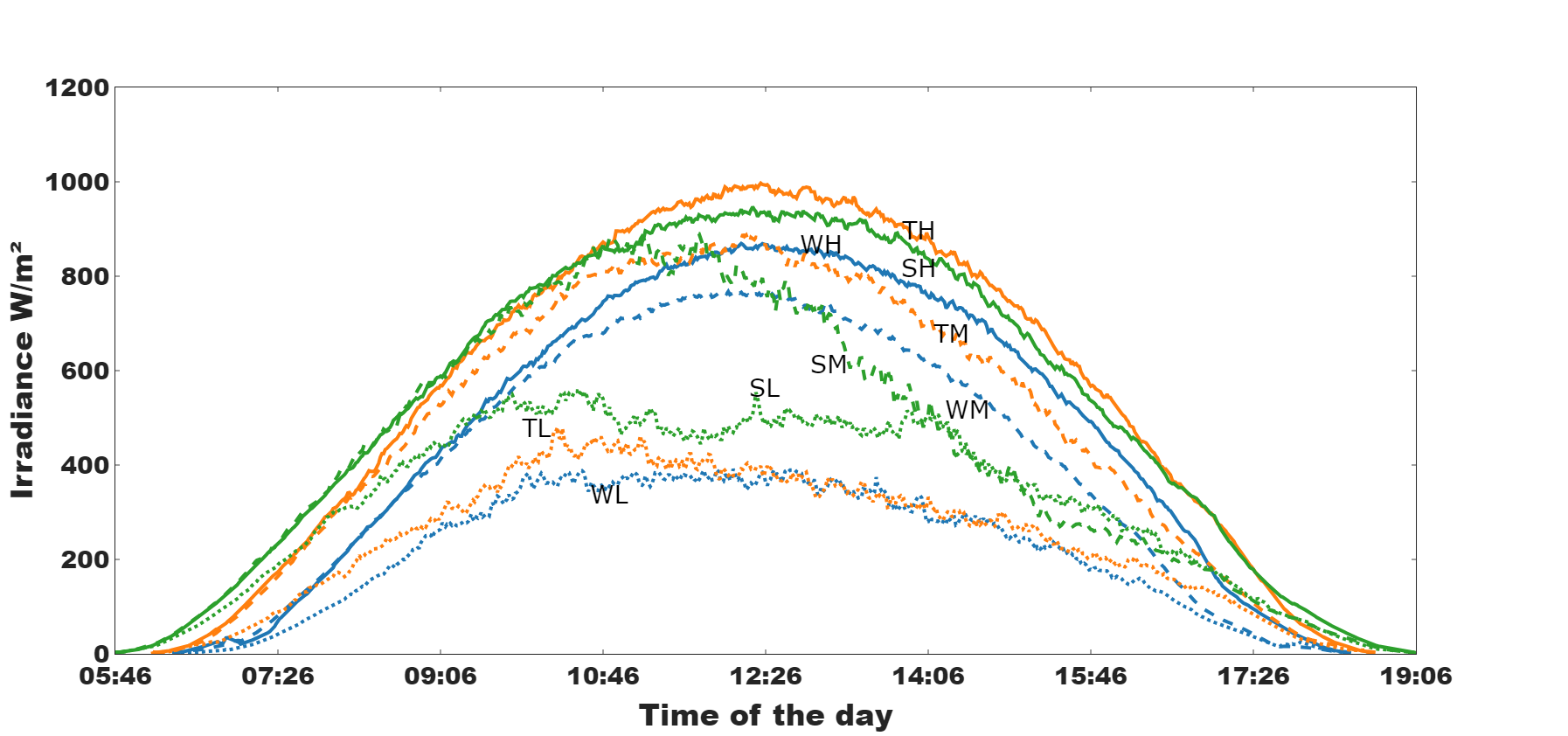}
			\caption{}
			\label{fig: ED mean profiles}
		\end{subfigure}
		\begin{subfigure}[b]{\columnwidth}
			\centering
			\includegraphics[width = 0.9\textwidth]{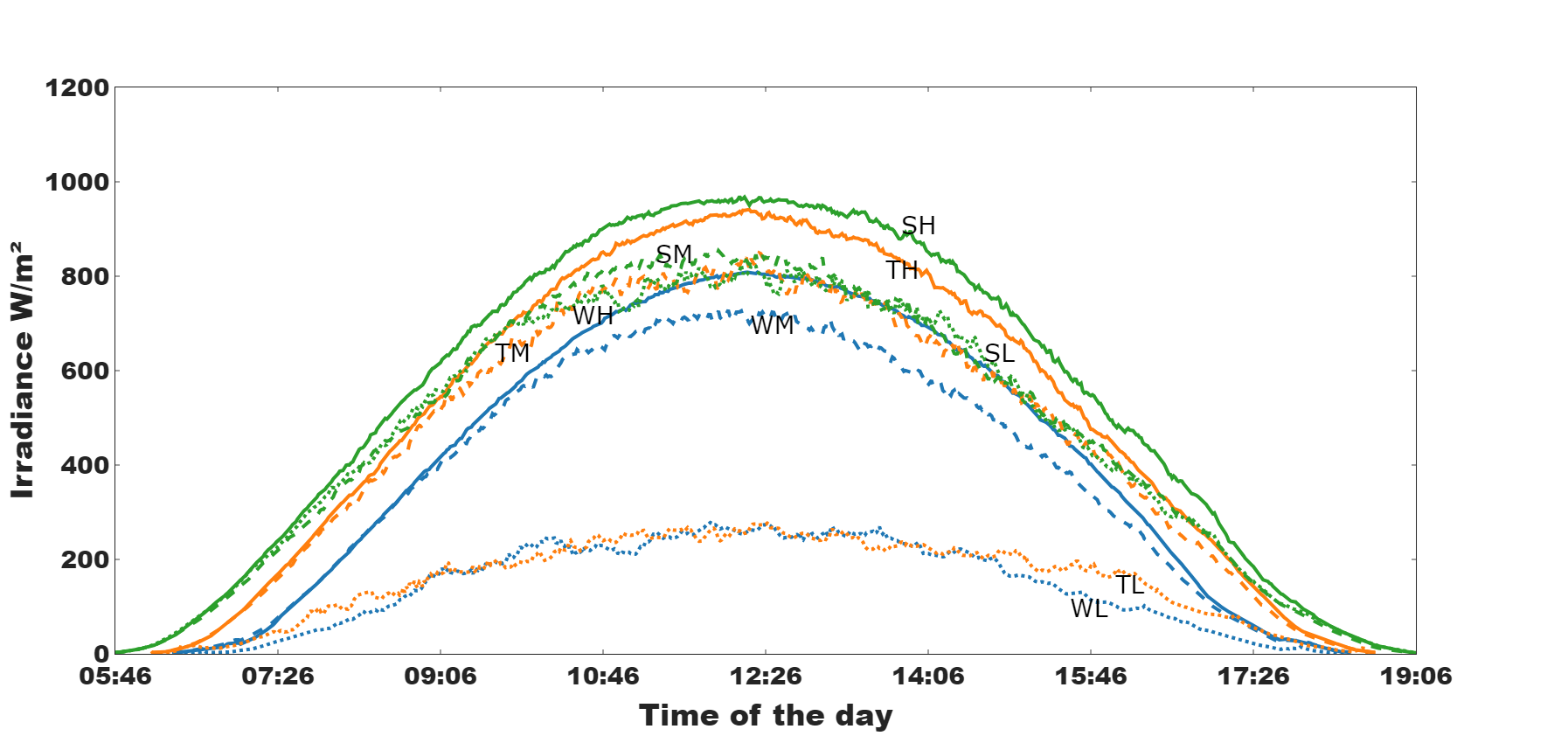}
			\caption{}
			\label{fig:DTW mean profiles}
		\end{subfigure}
		\caption{Cluster mean irradiance profiles obtained by different clustering approaches for Hawaii: (a) $\beta$-based, (b) ED time-series, and (c) DTW time-series. The first letter in the label represents the seasonal cluster (S, T or W), and the second letter represents the irradiance level cluster (H, M and L). The seasons S, T and W are shown in green, orange and blue lines, while the H, M, and L clusters in each season as solid, dashed, and dotted lines corresponding to clear, partially cloudy, and cloudy days, respectively.}
		\label{fig: mean profiles Hawaii}
	\end{figure}
}

\newcommand{\figureSeasonalSecondDayTransitionGoldenHawaii}{
	\begin{figure}[htbp]
		\centering
		\begin{subfigure}[b]{\columnwidth}
			\centering
			\includegraphics[width = \textwidth]{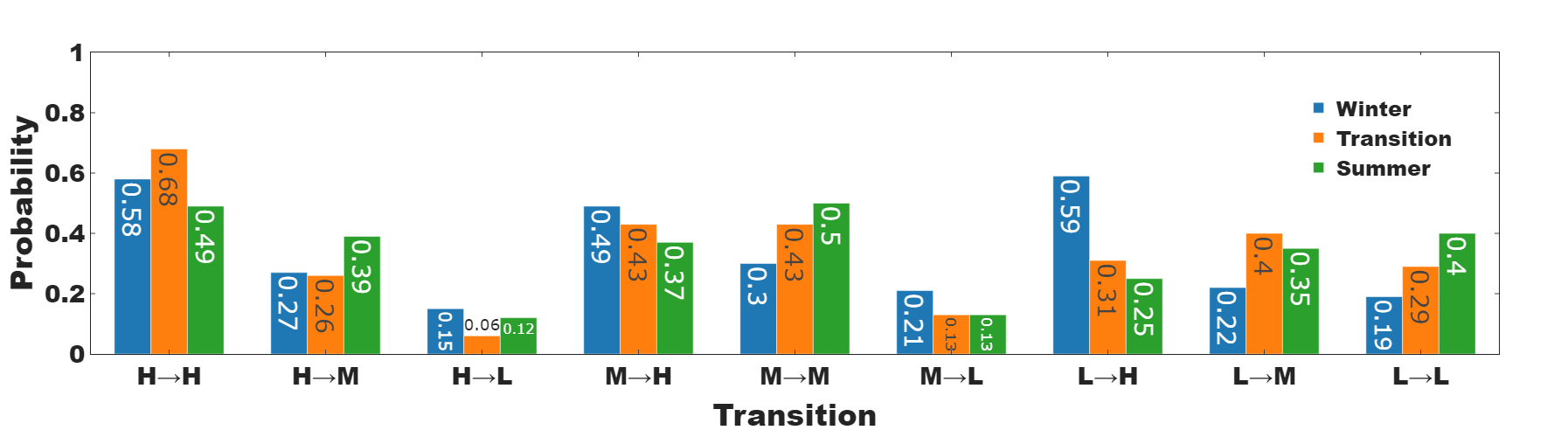}
			\caption{}
		\end{subfigure}
		\begin{subfigure}[b]{\columnwidth}
			\centering
			\includegraphics[width = \textwidth]{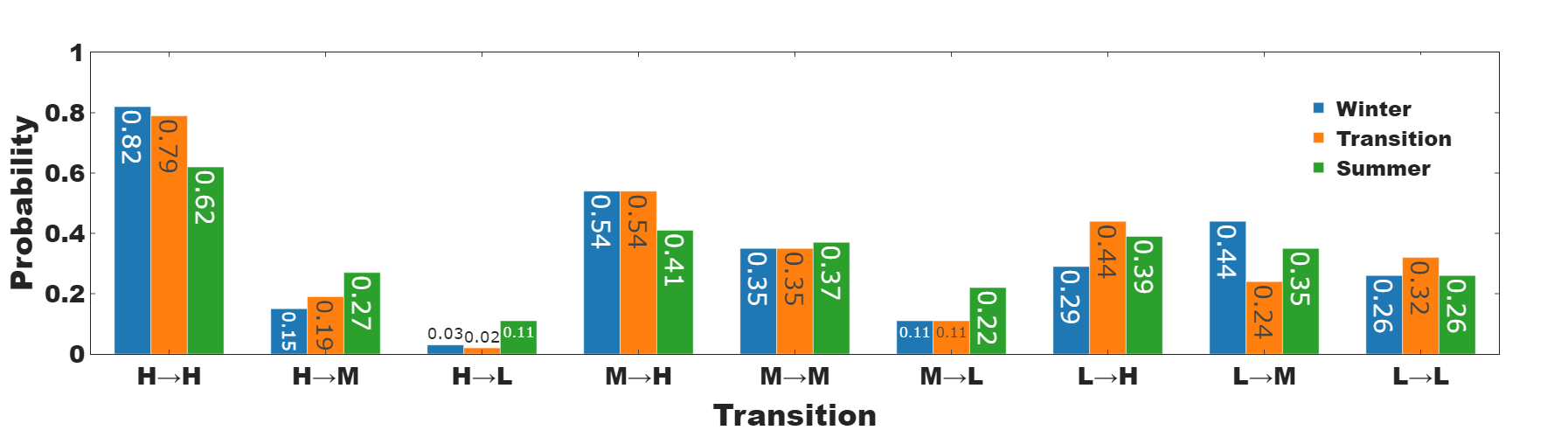}
			\caption{}
		\end{subfigure}
		\caption{Transition probabilities for different seasons obtained using $\beta$-based clustering approaches for: (a) Golden, and (b) Hawaii.}
		\label{fig: seasonal consecutive days transition Golden Hawaii}
	\end{figure}
}

\newcommand{\tableClusteringPerformanceGolden}{
	\begin{table}[htbp]
		\centering
		\caption{Clustering performance metrics of various approaches for Golden.}
		\resizebox{\linewidth}{!}{
			\begin{tabular}{cccccccc}
				\toprule
				
				\textbf{Approach} & \textbf{Dataset} & \multicolumn{3}{c}{\textbf{Clusters (No. of Days)}} & \textbf{Silh Score} &     \textbf{CH Score} &  \textbf{DB Score} \\
				& & \textbf{High} & \textbf{Medium} & \textbf{Low} & & & \\ 
				\midrule
				\textbf{ $\boldsymbol{\beta}$} &  \textbf{Winter} & 258 & 140 & 89 &    0.63 &  2102 &   0.48 \\
				& \textbf{Transition}   & 235 &  128 & 44 &    0.63 &  1401 &  0.48 \\
				& \textbf{Summer}  &  223 & 226  & 90 &   0.58 &  1619 &  0.51 \\
				& \textbf{All Season} & 720 & 528 & 212 & 0.60    &  4893 & 0.49   \\ \hline
				\textbf{ED} & \textbf{Winter}  		& 163 & 245 & 106  & 0.12 &  597 &   1.32 \\
				& \textbf{Transition}   &  153 & 197 & 57 &  0.24 &  456 &   1.01 \\
				& \textbf{Summer}  &  234 & 172 & 133 &   0.35 &  821 &  0.72 \\
				& \textbf{All Season}   & 610 & 515 & 335 & -0.07 &  120  &  2.92  \\ \hline
				\textbf{DTW}  & \textbf{Winter} &  205 & 235 & 74 & 0.14 &  348 &   1.85 \\
				& \textbf{Transition}   &  212 & 167 & 28 &  0.20   & 186  & 1.50   \\
				& \textbf{Summer}  &   104 & 311 & 124 &  0.20 &  196 &  1.27 \\
				& \textbf{All Season} & 436 & 534 & 490 & 0.03 &  108  &  6.09   \\
				\bottomrule
			\end{tabular}
		}
		\label{table: clustering performance score Golden}
	\end{table}
}

\newcommand{\tableClusteringPerformanceHawaii}{
	\begin{table}[htbp]
		\centering
		\caption{Clustering performance metrics of various approaches for Hawaii.}
		\resizebox{\linewidth}{!}{
			\begin{tabular}{cccccccc}
				\toprule
				\textbf{Approach} & \textbf{Dataset} & \multicolumn{3}{c}{\textbf{Clusters (No. of Days)}} & \textbf{Silh Score} &     \textbf{CH Score} &  \textbf{DB Score} \\
				& & \textbf{High} & \textbf{Medium} & \textbf{Low} & & & \\
				\midrule
				\textbf{ $\boldsymbol{\beta}$} &  \textbf{Winter} & 375 & 108 & 34  &    0.69 &  1812 &   0.47 \\
				& \textbf{Transition}   & 287 & 92 & 25 &    0.66 &  1349 &  0.46 \\
				& \textbf{Summer}  &  292 & 168  & 92 &   0.58 &  1395 &  0.53 \\
				& \textbf{All Season} & 878 & 451 & 131 & 0.61    &  3782 & 0.53   \\ \hline
				\textbf{ED} & \textbf{Winter}  & 135 & 322 & 60  & 0.02 &  569 &   1.21 \\
				& \textbf{Transition}   & 150 & 208 & 46 &  0.25 &  539 &   0.85 \\
				& \textbf{Summer}  &  381 & 84 & 74 &   0.39 &  657 &  0.74 \\
				& \textbf{All Season}   & 611 & 681 & 168 & 0.09 &  894  &  2.27  \\ \hline
				\textbf{DTW}  & \textbf{Winter} & 256 & 240 & 21 & 0.25 &  289 &   1.08 \\
				& \textbf{Transition}   &  239 & 144 & 21 &  0.27   & 305  & 1.13   \\
				& \textbf{Summer}  &   126 & 189 & 224 &  0.01 &  45 &  10.32 \\
				& \textbf{All Season} & 688 & 675 & 97 & 0.25 &  763  &  1.07   \\
				\bottomrule
			\end{tabular}
		}
		\label{table: clustering performance score Hawaii}
	\end{table}
}

\newcommand{\tableBetaRangeGolden}{
	\begin{table}[htbp]
		\centering
		\caption{The $\beta$ range observed for clusters obtained by different clustering approaches for Golden.}
		\label{table: beta range Golden}
		\resizebox{\linewidth}{!}{
			\begin{tabular}{cccccc}
				\toprule
				\textbf{Approach} & \textbf{Irradiance Level} &  \textbf{Winter} &     \textbf{Transition} &  \textbf{Summer} & \textbf{All Season}\\
				\midrule
				\textbf{$\boldsymbol{\beta}$} & \textbf{High}             &        [0.88, 1.19] & [0.83, 1.13] &  [0.8, 1.04] &  [0.84, 1.19]\\
				& \textbf{Medium}      &        (0.55, 0.88) &  (0.49, 0.83) &   (0.5, 0.8) &  (0.51, 0.84)\\
				& \textbf{Low} &       [0.1, 0.55] &  [0.07, 0.49] &  [0.1, 0.5] &  [0.07, 0.51]\\ \hline
				\textbf{ED} & \textbf{High}             &        [0.8, 1.13] & [0.76, 1.13] &  [0.66, 1.04] &  [0.71, 1.07]\\
				& \textbf{Medium}      &        [0.5, 1.19] &  [0.46, 1.09] &   [0.49, 0.88] & [0.47, 1.13] \\
				& \textbf{Low} &       [0.1, 0.75] &  [0.07, 0.68] &  [0.1, 0.72] &  [0.07, 1.19] \\ \hline
				\textbf{DTW} & \textbf{High}             &        [0.27, 1.13] & [0.24, 1.13] &  [0.47, 1.04] & [0.24, 1.19] \\
				& \textbf{Medium}      &        [0.31, 1.19] &  [0.37, 1.02] &   [0.38, 0.98] & [0.32, 1.07] \\
				& \textbf{Low} &       [0.1, 0.86] &  [0.07, 0.53] &  [0.1, 0.98] &  [0.07, 1.13]\\
				\bottomrule
			\end{tabular}
		}
	\end{table}
}

\newcommand{\tableBetaRangeHawaii}{
	\begin{table}[htbp]
		\centering
		\caption{The $\beta$ range observed for clusters obtained by different clustering approaches for Hawaii.}
		\label{table: beta range Hawaii}
		\resizebox{\linewidth}{!}{
			\begin{tabular}{cccccc}
				\toprule
				\textbf{Approach} & \textbf{Irradiance Level} &  \textbf{Winter} &     \textbf{Transition} &  \textbf{Summer} & \textbf{All Season}\\
				\midrule
				\textbf{$\boldsymbol{\beta}$} & \textbf{High}             &        [0.91, 1.11] & [0.87, 1.07] &  [0.86, 1.07] &  [0.89, 1.11]\\
				& \textbf{Medium}      &        (0.61, 0.91) &  (0.54, 0.86) &   (0.66, 0.86) &  (0.62, 0.89)\\
				& \textbf{Low} &       [0.10, 0.57] &  [0.04, 0.50] &  [0.05, 0.66] &  [0.04, 0.62]\\ \hline
				\textbf{ED} & \textbf{High}             &        [0.84, 1.11] & [0.84, 1.08] &  [0.67, 1.07] &  [0.80, 1.09]\\
				& \textbf{Medium}      &        [0.64, 1.10] &  [0.63, 1.09] &   [0.56, 0.86] & [0.56, 1.11] \\
				& \textbf{Low} &       [0.10, 0.78] &  [0.04, 0.73] &  [0.05, 0.72] &  [0.04, 0.86] \\ \hline
				\textbf{DTW} & \textbf{High}             &        [0.26, 1.11] & [0.35, 1.09] &  [0.45, 1.07] & [0.31, 1.11] \\
				& \textbf{Medium}      &        [0.43, 1.11] &  [0.48, 1.04] &   [0.05, 1.03] & [0.48, 1.07] \\
				& \textbf{Low} &       [0.10, 0.70] &  [0.04, 0.54] &  [0.49, 1.02] &  [0.04, 0.96]\\
				\bottomrule
			\end{tabular}
		}
	\end{table}
}

\newcommand{\tableConfusionMatrixGolden}{
	\begin{table}[htbp]
		\centering
		\caption{Normalized confusion matrix for ED-based and DTW distance-based clustering considering $\beta$-based clustering as the reference for Golden.}
		\label{tab: confusion matrix Golden}
		\resizebox{0.9\linewidth}{!}{
			\begin{tabular}{cc|ccc|ccc}
				\toprule
				\multirow{2}{4em}{\centering \textbf{Season}}&  \multirow{2}{5em}{\centering \textbf{Irradiance Level}}      & \multicolumn{3}{c|}{\textbf{Euclidean Distance}} & \multicolumn{3}{c}{\textbf{DTW Distance}} \\
				&        & \textbf{High} & \textbf{Medium} & \textbf{Low} & \textbf{High} & \textbf{Medium} & \textbf{Low}  \\
				\midrule
				\textbf{Winter} & \textbf{High} &   \textcolor{Green}{0.45} &  0.55 &   0 &   \textcolor{Green}{0.56} &  0.44 &   0 \\ 
				& \textbf{Medium} &    0.05 &  \textcolor{Green}{0.79} &  0.16 &     0.23 &  \textcolor{Green}{0.71} &   0.06 \\
				& \textbf{Low} &     0 &    0.07 &  \textcolor{Green}{0.93} &     0.11 &   0.16 &  \textcolor{Green}{0.73} \\
				\hline
				\textbf{Transition} & \textbf{High} &   \textcolor{Green}{0.65} &   0.35 &   0 &   \textcolor{Green}{0.70} &  0.30 &   0 \\
				& \textbf{Medium} &     0.05 &  \textcolor{Green}{0.84} &  0.11 &    0.25 &  \textcolor{Green}{0.73} &   0.02 \\
				& \textbf{Low} &     0 &    0.04 &  \textcolor{Green}{0.96} &    0.27 &   0.15 &  \textcolor{Green}{0.58} \\
				\hline
				\textbf{Summer} & \textbf{High} &   \textcolor{Green}{0.89} &   0.11 &   0 &   \textcolor{Green}{0.44}  &   0.45  & 0.11 \\
				& \textbf{Medium} &    0.16 &  \textcolor{Green}{0.65} &  0.19 &    0.02 &  \textcolor{Green}{0.84}  &  0.14 \\
				& \textbf{Low} &     0 &    0.02 &  \textcolor{Green}{0.98} &    0.01 &   0.24 &  \textcolor{Green}{0.75} \\
				\bottomrule
			\end{tabular}
		}
	\end{table}
}

\newcommand{\tableConfusionMatrixHawaii}{
	\begin{table}[htbp]
		\centering
		\caption{Normalized confusion matrix for ED-based and DTW distance-based clustering considering $\beta$-based clustering as the reference for Hawaii.}
		\label{tab: confusion matrix Hawaii}
		\resizebox{0.9\linewidth}{!}{
			\begin{tabular}{cc|ccc|ccc}
				\toprule
				\multirow{2}{4em}{\centering \textbf{Season}}&  \multirow{2}{5em}{\centering \textbf{Irradiance Level}}      & \multicolumn{3}{c|}{\textbf{Euclidean Distance}} & \multicolumn{3}{c}{\textbf{DTW Distance}} \\
				&        & \textbf{High} & \textbf{Medium} & \textbf{Low} & \textbf{High} & \textbf{Medium} & \textbf{Low}  \\
				\midrule
				\textbf{Winter} & \textbf{High} &   \textcolor{Green}{0.35} &  0.65 &   0 &   \textcolor{Green}{0.63} &  0.37 &   0 \\ 
				& \textbf{Medium} &    0.03 &  \textcolor{Green}{0.73} &  0.24 &     0.12 &  \textcolor{Green}{0.24} &   0.59 \\
				& \textbf{Low} &     0 &    0 &  \textcolor{Green}{1} &     0.18 &   0.24 &  \textcolor{Green}{0.59} \\
				\hline
				\textbf{Transition} & \textbf{High} &   \textcolor{Green}{0.52} &   0.48 &   0 &  \textcolor{Green}{0.73} &  0.27 &   0 \\
				& \textbf{Medium} &     0.01 &  \textcolor{Green}{0.76} &  0.23 &    0.27 &  \textcolor{Green}{0.72} &   0.01 \\
				& \textbf{Low} &     0 &    0 &  \textcolor{Green}{1} &    0.16 &   0.04 &  \textcolor{Green}{0.80} \\
				\hline
				\textbf{Summer} & \textbf{High} &   \textcolor{Green}{1} &   0 &   0 &   \textcolor{Green}{0.38}  &   0.35  & 0.27 \\
				& \textbf{Medium} &    0.53 &  \textcolor{Green}{0.40} &  0.07 &    0.03 &  \textcolor{Green}{0.30}  &  0.67 \\
				& \textbf{Low} &     0 &    0.2 &  \textcolor{Green}{0.80} &    0.11 &   0.47 &  \textcolor{Green}{0.42} \\
				\bottomrule
			\end{tabular}
		}
	\end{table}
}

\begin{document}

\begin{frontmatter}



\title{\colorOpt{red}{Two-level Solar Irradiance Clustering with Season Identification: A Comparative Analysis}}


\author[a]{Roshni Agrawal}
\author[a]{Sivakumar Subramanian \corref{1}}
\author[a]{Venkataramana Runkana}
\affiliation[a]{organization={Tata Research Development and Design Centre},
            addressline={Tata Consultancy Services Limited}, 
            city={Pune},
            country={India}}

\cortext[1]{Corresponding author. 
Email address: sivakumar.subramanian1 at tcs.com (Sivakumar Subramanian)}

\begin{abstract}
\colorOpt{red}{
Solar irradiance clustering can enhance solar power capacity planning and help improve forecasting models by identifying similar irradiance patterns influenced by seasonal and weather changes. In this study, we adopt an efficient two-level clustering approach to automatically identify seasons using the clear sky irradiance in first level and subsequently to identify daily cloud level as clear, cloudy and partly cloudy within each season in second level. In the second level of clustering, three methods are compared, namely, Daily Irradiance Index (DII or $\beta$), Euclidean Distance (ED), and Dynamic Time Warping (DTW) distance. The DII is computed as the ratio of time integral of measured irradiance to time integral of the clear sky irradiance. The identified clusters were compared quantitatively using established clustering metrics and qualitatively by comparing the mean irradiance profiles. The results clearly establish the superiority of the $\beta$-based clustering approach as the leader, setting a new benchmark for solar irradiance clustering studies. Moreover, $\beta$-based clustering remains effective even for annual data unlike the time-series methods which suffer significant performance degradation. Interestingly, contrary to expectations, ED-based clustering outperforms the more compute-intensive DTW distance-based clustering. The method has been rigorously validated using data from two distinct US locations, demonstrating robust scalability for larger datasets and potential applicability for other locations.
}
\end{abstract}


\begin{highlights}

\colorOpt{red}{\item Robust irradiance clustering approach tested for two locations in the US.
\item Seasonal boundaries clearly identified using clear sky irradiance data.
\item Daily Irradiance Index (DII) efficiently characterizes solar irradiance levels.
\item DII-based clustering outperforms tested time-series based methods.
\item Irradiance level transitions for each season analyzed.
}
\end{highlights}


\end{frontmatter}


\begin{singlespacing}
\section*{Acronyms}
\begin{acronym}[AAAAA]\acro{BMS}{Baseline Measurement System}
\acro{CC}{Cloud Cover}
\acro{CH}{ Calinski-Harabasz} 
\acro{CSI}{Clear Sky Irradiance}
\acro{CSR}{Clear Sky Ratio}
\acro{DB}{Davies-Bouldin}
 \acro{DII}{Daily Irradiance Index} 
\acro{DTW}{Dynamic Time Warping}
\acro{ED}{Euclidean Distance}
\acro{GHI}{Global Horizontal Irradiance}
\acro{H}{High}
\acro{L}{Low}
\acro{MD}{Manhattan Distance}
\acro{M}{Medium}
\acro{NELHA}{Natural Energy Laboratory of Hawaii Authority}
\acro{NREL}{National Renewable Energy Laboratory}
\acro{SOM}{Self-organizing Maps}
\acro{Silh}{Silhouette} 
\acro{SRRL}{Solar Radiation Research Laboratory}
\acro{S}{Summer}
\acro{SH}{Summer High}
\acro{SM}{Summer Medium}
\acro{SL}{Summer Low}
\acro{T}{Transition}
\acro{TH}{Transition High}
\acro{TM}{Transition Medium}
\acro{TL}{Transition Low}
\acro{W}{Winter}
\acro{WH}{Winter High}
\acro{WM}{Winter Medium}
\acro{WL}{Winter Low}

\end{acronym}
\end{singlespacing}

\section{Introduction}
Many countries are investing substantially in renewable energy sources in response to urgent calls for combating climate change. Particularly, solar photovoltaics (PV) has been witnessing significant growth in the last decade. The world's installed solar capacity has surged almost 10 times in the last decade, going from 104 GW in 2012 to 1053 GW in 2022 \citep{Ourworld}. Geographies that are leading the chart are China (393 GW), US (113 GW), Japan (78 GW), Germany (66 GW) and India (63 GW).  The increased integration of PV systems into the grid presents challenges for grid stability due to dynamic changes in PV power generation. To alleviate the problem, reliable forecast of solar generation is crucial. This underscores the necessity for robust solar power forecasting models to support grid planning and scheduling.

The instantaneous power generation of a given PV system depends primarily on the irradiance received by solar panels and ambient (module) temperature. While solar irradiance reaching the surface of earth goes through predictable diurnal and seasonal variations, its variation on any given day is uncertain due to various atmospheric phenomena. They influence observable factors such as clouds (type and intensity), wind velocity, temperature, and relative humidity, these in turn affect the transmission and scattering of solar radiation. Developing accurate solar forecasting models is challenging, primarily due to the complexity associated with characterizing these atmospheric influences. The increasing demand for better forecasting has led to a variety of solutions. Researchers have employed diverse techniques, including time-series prediction with statistical methods, machine learning and deep learning algorithms \citep{kim2023}, as well as atmospheric modeling using sky images and cloud dynamics \citep{paletta2021}. These models rely on meteorological and/or historical power generation data. In this context, the clustering of solar irradiance data is one of the earliest techniques used to group similar irradiance profiles to aid the development of forecasting models \citep{chun2017b, baor2018, paul2018, hui2018, spyr2020, maoYang2022}. 

Irradiance clustering can offer a quantitative assessment of solar power generation potential at a site. It could be used to estimate daily, seasonal, or yearly yields \citep{paul2019, isab2021}. Moreover, many researchers have presented the application of irradiance clustering for site selection, design of solar plants, sizing storage facilities and electric loads, and to reliably project annual performance using solar irradiance data collected from the location \citep{fahd2020}. For example, a case study delved into the sizing of a standalone PV system integrated with an anaerobic digestion biogas power plant \citep{chun2017b}. Another study explored the relationship between the sizing of an electric vehicle charger and storage facility with solar irradiation characteristics \citep{josh2013}. This brief review and discussion demonstrate diverse applications of clustering in addressing distinct solar energy-related challenges.

Irradiance clustering techniques can be broadly categorized into two types: time-series distance-based clustering and feature-based clustering. In the widely used time-series distance-based clustering, time-series distances between all the pairs of daily profiles are calculated to group similar days. The time-series profiles may be of measured irradiance or of derived indices such as clearness index, and clear
sky index \citep{lauret2022}.  On the other hand, in feature-based clustering, the clustering is performed on one or many features extracted from daily time-series profiles. Features, such as total irradiance, mean irradiance, and peak irradiance, can serve as representative attributes for a day.

Time-series similarity matrix for $N$ daily profiles to be clustered shall be of dimension $N\times N$. The distances between two time-series may be measured using Euclidean Distance (ED), Dynamic Time Warping (DTW) distance or Manhattan Distance (MD). ED and MD have point-to-point mapping whereas DTW has many-to-one point mapping for optimal alignment between two time-series profiles, calculating minimum distance for the warping path. The key drawback of time-series approach is the computational complexity associated with computing the similarity matrix for a large dataset. This is more pronounced for DTW metric as its estimation involves solving a path optimization problem that is sensitive to the sunshine duration of daily profiles. Time-series clustering, including ED \citep{chun2017b, zhang2021, cong2019a}, DTW \citep{isab2021, chun2017b, chun2017a, ebuka2021, tanaka2016} and MD \citep{chun2017b} have been used for irradiance clustering. The popularity of DTW distance-based time-series clustering is attributed to its flexibility in accommodating varying sunshine duration of days observed in various seasons \citep{chun2017b}.

In the second approach, features derived from daily irradiance profiles along with representative weather-related parameters are used for clustering. This approach results in a more modest feature matrix of dimension $N \times M$, where $N$ is the number of days and $M$ is the number of features to be used for clustering, typically, $M \ll N$. Clearly, the performance of the approach is sensitive to the features used in the clustering. This approach was used for spatial clustering to group geographies with similar solar generation potential in \citep{take2016, maldonado2022}. A study used five features mean, standard deviation, skewness, kurtosis, and moving fluctuation intensity \citep{xing2018}, and another used six features derived from Clear Sky Ratio (CSR) namely dusky sunshine duration, cloudy sunshine duration, average CSR under cloudy, standard deviation of CSR under cloudy, maximum change rate of CSR under cloudy, and average change rate of CSR under cloudy \citep{shua2017}. 

Additionally, irradiance clustering approaches can be further classified into single-level and two-level clustering. In single-level clustering, the days are clustered independent of the seasons into different groups either using a time-series distance-based similarity metrics \citep{paul2019, baor2018, patr2013} or using one or many features of irradiance profiles \citep{take2016, cong2019a, azed2019}. This could result in ambiguous clusters grouping days from different seasons but with similar solar yield or irradiance pattern.

To bring in the seasonal distinction, many authors have recommended two-level clustering \citep{chun2017b, chun2017a, xing2018, hui2018}. The first level is designed to identify the seasons while the second level is expected to cluster days of similar irradiance levels within a season. As only the days of the same season are clustered in the second level, this approach has the advantage of reducing the computational burden especially when used with metrics such as DTW distance. Two-level clustering may be applied without considering seasons as in \cite{isab2021}. In the first level, they used total daily irradiance to identify three groups. In the second level, they used DTW time-series based clustering on the variability time-series profiles constructed by differences between two successive instantaneous irradiance values.

Some authors took a much simpler approach for identifying the seasons with calendar-based rules \citep{chun2017b, chun2017a, xing2018, hui2018}. While this approach will work well for a specific location but requires customization for other locations or regions. Therefore, automating the identification of seasonal boundaries using clustering methods would be preferred as it offers greater flexibility and adaptability, improving clustering performance and eliminating the reliance on regional calendar rules for season identification \citep{rosh2023}. Moreover, irradiance forecasting models should be updated regularly to adapt to varying seasonal and weather conditions for accurate prediction. Seasonal clustering can help construct a small number of models representing each season, reducing the amount of calculation in weather updates \citep{tanaka2016}.

The clustering algorithm used in these studies also played an important role in identifying the clusters. K-Means clustering algorithm stands out as the most widely used algorithm for solar irradiance clustering \citep{chun2017b, paul2018, baor2018, hui2018, ribeiro2024, cong2019a}. K-Means++ is a similar algorithm, but it improves results by randomizing the initialization to distribute initial centroids far apart to prevent convergence to suboptimal minima \citep{isab2021}. Both K-Means and K-Means++ utilize the mean to represent cluster centers, making them susceptible to outliers such as non-prevalent weather conditions experienced in a location. K-Medoids clustering algorithm overcomes this limitation by using a medoid to represent each cluster. Unlike the mean in K-Means, the medoid is an actual data point that is centrally located within the cluster, ensuring a more representative choice. For example, K-Means clustering algorithm was used to cluster the daily temporal cloud cover in \citep{paul2019} and daily direct fraction profiles in \citep{patr2013}. K-Medoids algorithm was used to cluster clear sky direct normal irradiance profiles in \cite{baor2018} and daily irradiance profiles in \citep{ribeiro2024, cong2019a, chun2017b}. Some of the researchers have also adopted methods such as self-organizing maps (SOM) \citep{take2021, javier2022} for performing feature-based clustering on solar irradiance data. 

\colorOpt{red}{Several clustering methods were investigated to determine the optimum number of clusters presented in a photovoltaic power patterns dataset from a single season in \cite{MuMo2016}. The analysis included K-means from partitional clustering, Hierarchical from agglomerative clustering, Fuzzy C-means from fuzzy clustering, Self-Organizing Maps from neural network-based algorithms, and Ant Colony and Bat from bio-inspired optimization methods. Interestingly, multiple clustering quality indices Davies Bouldin, Dunn, Silhouette, Bayesian Information Criterion, were used to track the performance of the clustering methods and to determine the number of clusters. Further details of the clustering methods and performance metrics can be found in the referenced source \cite{MuMo2016}. They reported that bio-inspired bat method was the most efficient and outperformed the other clustering methods. Interestingly, even for a single season studied, the algorithms recommended more than seven or eight clusters. In comparison, in this work, we seek only three clusters, corresponding to clear, partly cloudy and cloudy days, for each season.
}

\colorOpt{red}{
Recently, we addressed the critical gaps in existing solar irradiance clustering studies and proposed an elegant two-level clustering of solar irradiance \citep{rosh2023} with automated seasons identification utilizing only the features from clear sky irradiance in the first level.} This approach clearly identified contiguous seasons in a year. In the second level, feature-based clustering with daily irradiance index (\ac{DII} or $\beta$) was adopted for clustering days into clear, partly cloudy and cloudy days. The $\beta$-based method is computationally efficient as it uses a single physically significant measure for clustering. It is worth noting that prior studies \citep{chris2013, hui2018, bali2020} have utilized a measure similar to $\beta$. \cite{chris2013, bali2020} used empirical rules to classify days into distinct categories such as clear, partially cloudy and cloudy, etc. and  in \cite{hui2018} the focus was to train different forecasting models for different clusters and not in evaluating the efficacy of the clustering approach. Most importantly, past works have not compared $\beta$-based method to other known clustering approaches. In this work, we endeavor to fill these important gaps establishing the advantages and significance of the approach.

\colorOpt{red}{The aim of the work is to strengthen our hypothesis for leveraging $\beta$-based approaches for advancing solar energy analytics. The specific contributions of the study are: a) providing a direct comparison of $\beta$ to the daily average cloud cover ascertaining that it is in alignment to the sky-image based cloud cover measurements,  b) establishing that $\beta$-based clustering identifies well-defined unambiguous clusters when compared to the tested time-series distance-based methods, c)  demonstrating that $\beta$-based clustering works well for both seasonal and annual data unlike time-series methods whose performance degrades significantly if applied for the annual data,  d) contrary to the general expectations, highlighting that the simple ED-based approach works better than computationally complex DTW distance-based clustering, and e) providing a probabilistic analysis of irradiance level transitions from day to day for each season that can be used for simulation and solar forecasting studies. The subject of the research is to validate the robustness, efficiency, and generalization of the $\beta$-based approach taking two distinct representative geographical locations, namely Hawaii and Golden, Colorado.}

\colorOpt{red}{The paper is organized as follows. In Section \ref{proposed approach}, the two-level clustering approach and metrics for clustering assessment are discussed in detail. Section \ref{sec:irrdatabase} introduces the irradiance database used in the study. A comparison of clustering results from the popular ED-based and DTW distance-based time-series clustering to $\beta$-based clustering is presented in Section \ref{results} and Section \ref{conclusion} concludes the paper.}

\section{Methodology}\label{proposed approach}
In this work, as shown in Figure \ref{fig:twoLevelFlowChart}, a two-level clustering framework proposed by \citep{rosh2023} is adopted. The first level is used to identify seasonal boundaries using features extracted from the clear sky irradiance (CSI). In particular, the total CSI and sunshine duration were found to be sufficient in this level.

\figureTwoLevelFlowChart

The second level of clustering focuses on detecting variability within a season using an integral feature extracted from the measured irradiance referred to as Daily Irradiance Index (DII also referred as  $\beta$). It is a normalized scalar index that characterizes the daily measured irradiance level compared to daily clear sky irradiance level, defined as:
\begin{equation}
\beta = \frac{\text{Area under the daily measured irradiance}}{\text{Area under the daily clear sky irradiance}}
\end{equation}

By comparing the measured irradiance profile with the clear sky irradiance profile, $\beta$ quantifies the effect of atmospheric factors on solar irradiation at the day level. Its value typically ranges from 0 to 1, where the extremes represent a fully cloudy day and a clear day, respectively. It has been reported that the upper limit of 1 represents a soft bound with some days reporting more irradiance levels than expected from the clear sky model \citep{isab2021}.

As $\beta$ is computed independently for each day like any other feature, clustering becomes rather straightforward unlike time-series ED or DTW distances discussed earlier that require pair-wise distances. In this work, K-Means clustering algorithm is used in both levels for its simplicity. Further, the features extracted from the dataset were checked to be free of outliers.

As shown in the Figure \ref{fig:twoLevelFlowChart}, this work further endeavors to compare the clustering results from the popular ED and DTW time series distance-based clustering in the second level 
to $\beta$-based clustering, taking the latter to be the reference. This is justified considering the directness of the measure in capturing the irradiance level of any day. \colorOpt{red}{To reinforce this idea further, a direct or single-level clustering for all seasons is also performed.} Note that for ED and DTW distance-based time-series clustering, K-Medoid clustering is utilized due to its robustness against noise and outliers present in dataset.

\subsection{Cluster Quality Measures}
The identified clusters from all the methods are first assessed independently using three metrics, namely Silhouette (\ac{Silh}) score, Calinski-Harabasz (\ac{CH}) score, and Davies-Bouldin (\ac{DB}) score \citep{scikit-learn}. These were evaluated using \textit{sklearn.metrics.cluster} submodule of \textit{scikit-learn} library \citep{scikit-learn}.

The Silhouette score is a measure that considers both clustering cohesion and separation. It is computed as the average of Silhouette coefficients for all days. These coefficients capture the distance between a daily observation and other observations within the same cluster, as well as the distance between the daily observation and observations in the nearest neighboring cluster. The Silh score varies between -1 and 1, a score closer to 1 indicates well-defined and distinct clusters.

The CH score is also known as the variance ratio criterion. It quantifies the ratio of the sum of inter-cluster dispersion and intra-cluster dispersion for all clusters. Dispersion is defined as the sum of squared distances. Larger CH scores suggest well-defined clusters with dense and well-separated clusters.

The DB score represents the average similarity measure of each cluster with its most similar cluster. Similarity is computed as the ratio of within-cluster distances to between-cluster distances. Clusters with better separation have lower similarity and yield a DB score close to zero. A lower DB score indicates better clusters.

In summary, better clustering results are indicated by a Silh score closer to one, a larger CH score, and a DB score closer to zero. These metrics provide a comprehensive evaluation of the clustering performance, enabling us to identify and recommend the most effective clustering approach for solar irradiance analysis. Prior studies \citep{rosh2023, de2018, ebuka2021, ribeiro2024} have used Silh score, CH score and DB score to validate the clusters.

\section{Irradiance Database}\label{sec:irrdatabase}
\colorOpt{red}{Details of the datasets used in the present work are provided in this section. The daytime measured irradiance and cloud cover data used in this study was sourced from the Baseline Measurement System (BMS) located at the National Renewable Energy Laboratory (NREL) Solar Radiation Research Laboratory (SRRL) in Golden, Colorado (latitude: 39.74, longitude: -105.18, elevation: 1828.8 meters above sea level). For our analysis, 1-minute sampled Global Horizontal Irradiance (GHI) and Total Cloud Cover (CC) data from January 2015 to December 2018 was considered \citep{nrel}. The GHI data was collected using the Global CMP 22 equipped with a Kipp \& Zonen pyranometer. The Total Cloud Cover was measured using Yankee Total Sky Imager Model 880. This instrument captures all sky images, computes percentage cloud cover and provides cloud decision images.} 

\colorOpt{red}{
We have also utilized a dataset from Natural Energy Laboratory of Hawaii Authority (NELHA) Hawaii Ocean Science \& Technology Park located in Kailua-Kona, Hawaii (latitude: 19.72, longitude: -156.05, elevation: 4 meters above sea level) to validate and further assess the robustness and generalizability of our approach. We have considered 1-min sampled GHI measured by a Kipp \& Zonen Pyranometer from January 2015 to December 2018 \citep{nrelnelha}. Outliers and missing records in the daytime data of GHI dataset were imputed by linear interpolation. Note that the cloud cover data was not available for this location. 

For both the locations in the study, the clear sky irradiance is estimated using Ineichen and Perez clear sky model \citep{ineichen2008} available in PVLIB Toolbox  as \textit{Location.get\_clearsky(times)} method that requires the geographical location and date range as inputs \citep{pvlib}.
}

\colorOpt{red}{
Figure \ref{fig:cloudcoversamples}(a) illustrates an example of a day experienced in Golden.} The instantaneous measured irradiance, clear sky irradiance, and total cloud cover (on the right y-axis) are shown for 17 February 2017. The corresponding all-sky images at different times of the day are presented in Figure \ref{fig:cloudcoversamples}(b)-(e). In these figures, the left image depicts raw capture from the sky camera, while the right image shows the corresponding cloud decision image using cloud cover retrieval processing, providing percentages of thin and opaque cloud cover estimates. In the cloud decision images, clear sky, opaque or thick clouds, thin clouds, masked camera arm and sun-blocking strip, are represented by colors medium blue, white, light blue, and black, respectively. Thick and thin clouds were identified using opacity classification algorithm \citep{nrel}. The instantaneous total cloud cover, later compared in this study, is the sum of the percentages of thin and thick cloud covers. A simple procedure was adopted to preprocess the cloud cover data to obtain daily daytime averages. Very few of the minute wise cloud cover data was found to contain non-physical negative values. These were replaced with Nan (not a number) entries to avoid using them in the average calculations. Days with less than 90 percent of the cloud cover data availability were dropped from the analysis. As the data availability was generally good, this resulted in the elimination of only six days from the four years of data.

\figureCloudCoverSamples

Examining cloud cover images at 8:00, 10:30, 13:00, and 15:30 in the Figures \ref{fig:cloudcoversamples}(b)-(e), we observe distinct irradiance patterns. At 8:00, an overcast sky with a total cloud cover of 93\% resulted in significantly lower measured irradiance compared to the clear sky irradiance. By 10:30, the sky cleared, with a total cloud cover of 13\%, leading to measured irradiance approaching clear sky levels. At 13:00, the sky was clear aligning measured irradiance with clear sky values. Later in the day, at 15:30, an overcast sky with a total cloud cover of 79\% led to lower measured irradiance.

\section{Results and Discussion}\label{results}
\colorOpt{red}{
The two-level clustering method was applied to data gathered from two locations from the US as explained in the previous section. These locations were primarily selected to represent diverse weather patterns and to help establish the robustness and generalizability of the proposed method.

}


\subsection{Level-1: Identification of Seasons }
\colorOpt{red}{In this level, features derived from clear sky irradiance (CSI) was used to identify seasons. In particular, sunshine duration and area under CSI were found to be sufficient to clearly identify the seasons. The sunshine duration varied from 560 to 896 minutes and the area under the CSI varied from 2.73 to 9.04 ${\rm kWh/m}^2$, respectively, for Golden. For Hawaii, the sunshine duration ranged from 654 to 797 minutes and the area under the CSI  varied from 4.69 to 7.75 ${\rm kWh/m}^2$. Due to Hawaii's lower latitude, sunshine duration remains relatively stable throughout the year. In contrast, Golden experiences more pronounced variations in sunshine duration because of its higher latitude and the greater influence of the Earth's axial tilt. 
}	

\figureSeasonsHawaiiSubplots

\colorOpt{red}{Four continguous repeating seasons of Winter (W), Transition (T), Summer (S) and Transition (T) were identified as shown in Figure \ref{fig: season scatter plot Hawaii and seasonal boundary with vertical colorbar}a and color coded scatter plot of CSI features used for identifying the seasons is shown in Figure \ref{fig: season scatter plot Hawaii and seasonal boundary with vertical colorbar}b for Hawaii. The seasonal clustering demonstrates strong performance across these seasons, achieving a Silh score, a CH score and a DB score of 0.63, 8080, and 0.46 for Hawaii, respectively. In an earlier study \cite{rosh2023}, we identified clear seasonal boundaries for Golden taking a similar approach. This emphasizes that the presented method is robust and works well for different locations.
}


\subsection{Efficacy of DII ($\beta$)}

A strong correlation between daily cloud cover and daily $\beta$ was expected which was evident from the sample profiles shown in Section \ref{sec:irrdatabase}. The daily $\beta$ was calculated using respective historical measured irradiance and clear sky irradiance for each day. For Golden, the daily mean of total cloud cover in the dataset is compared to the $\beta$ of the respective day in Figure \ref{fig:betasdistribution}. While there is a significant scatter in the data, the broad trend of the correlation is clear. The scatter may be due to the measurement errors both in the irradiance and cloud cover data. Moreover, cloud cover image processing used images of the entire horizon, thus representing a larger area. In comparison, irradiance measurements may have a strong local component. It has to be kept in mind that the cloud cover average puts equal emphasis on cloud level measurements gathered throughout the day. But $\beta$ may be more influenced by the data around solar noon as the expected irradiance levels are higher around this  time.  From the Figure \ref{fig:betasdistribution}, it is clear that $\beta$ is capturing the irradiance level of the day quite well. Further, two linear segments with distinct slopes are discernible. The first segment with a gradual slope covers the higher $\beta$ range of more than 0.6 with a corresponding cloud cover of 0-70\%. When the cloud cover increases further, it seems to bring down the $\beta$ sharply. The scatter in the plot perhaps emphasizes the complex interaction between cloud types, depth and coverage on the irradiance realized. 

\figureBetaCloudCover

The measured irradiance, clear sky irradiance, and cloud cover (on the right y-axis) for selected days representing low, medium and high cloud cover days are plotted in Figure \ref{fig:betasamples}. These days are shown as pink, orange, and green-filled circles, respectively, in the Figure \ref{fig:betasdistribution}.  The clear day trends shown in Figure \ref{fig:betasamplesHighBeta} with $\beta$ values close to 1 are as expected. They demonstrate minimal cloud activities with irradiance measurements mirroring or exceeding clear sky irradiance. Two of the four days shown had minor cloud activities during some parts of the day. Three of the four medium cloud cover days shown in Figure \ref{fig:betasamplesMedBeta} demonstrated highly variable cloud activities resulting in moderate $\beta$. Three of the four sampled high cloud cover days shown in Figure \ref{fig:betasamplesLowBeta} demonstrate significant cloud presence during most of the day resulting in poor irradiance with $\beta$ values lower than 0.5. These days are associated with overcast conditions, representing fully cloudy or mostly cloudy days. Qualitatively, the negative correlation between measured irradiance and cloud cover is well brought out in these figures. The observations support the hypothesis that $\beta$ serves as an effective measure for gauging the extent of irradiance on any given day.

\figureSampledIrrAndCloudCover

\colorOpt{red}{The Box and Whisker plots of $\beta$ distribution in a month for four years are shown in Figures \ref{fig:betaMonthlyGoldenHawaii}a and \ref{fig:betaMonthlyGoldenHawaii}b for Golden and Hawaii, respectively. It can be observed that the monthly median $\beta$ values change in a relatively narrow band for most of the months, and the $\beta$ spread is uneven around the mean. For example, in Figure \ref{fig:betaMonthlyGoldenHawaii}a, Jan 2015, May 2018, Oct 2018, and Nov 2018 show significant spread. Relatively speaking, July and August seemed to be consistent months with narrower spreads. Interestingly, the month of May stands out with the lowest mean $\beta$ value, perhaps attributed to increased cloud activity during this period. This plot affirms that significant cloud activity could be observed in any of the months (or seasons).
}

\figureBetaMonthlyYearOnYearGoldenHawaii

\colorOpt{red}{
The $\beta$ distribution in Figure \ref{fig:betaMonthlyGoldenHawaii}b  for Hawaii indicates that it has a stable climate year-around compared to Golden with most of days reporting $\beta >  0.6$. Overall, the tropical climate of Hawaii yielded consistently higher mean $\beta$ when compared to Golden.  It can be observed that the $\beta$ distributions of several months of the year 2018 showed significant deviations from the previous years.
}

\subsection{Level-2: Clustering with $\beta$, and time-series distance methods}
In the second level, feature-based clustering was performed using $\beta$. The resulting clusters are designated as High (H), Medium (M) and Low (L) based on the range of $\beta$ values, denoting clear, intermittent, and cloudy days, respectively. From the literature review discussed earlier, it is evident that time-series clustering based on ED and DTW distances have been popular. Results from these two methods are compared to the $\beta$-based clusters taking the latter to be the reference. This can be justified by the preceding discussion on the efficacy of $\beta$ as the reliable irradiance metric and for its ability to identify unambiguous clusters as we shall discuss here.

\colorOpt{red}{Figures \ref{fig: mean profiles Golden}a and \ref{fig: mean profiles Hawaii}a show the mean irradiance profiles of clusters for the three seasons obtained using $\beta$-based clustering for Golden and Hawaii, respectively. These mean profiles were computed by averaging the irradiance levels of all the days within a cluster for every sampling instant (minute). The labels in these plots adhere to a specific convention, where the first letter represents the seasonal cluster -- W, T and S, and the second letter represents the irradiance level cluster -- H, M and L. It can be observed that all the nine mean irradiance profiles are distinct, indicating well-separated clusters for both the locations.
}

\figureBetaEDDTWMeanProfilesGolden
\figureBetaEDDTWMeanProfilesHawaii

\colorOpt{red}{Similar cluster mean profiles obtained using ED-based and DTW distance-based clustering are shown in Figures \ref{fig: mean profiles Golden}b and Figure \ref{fig: mean profiles Golden}c for Golden. The corresponding profiles for Hawaii are presented in Figure \ref{fig: mean profiles Hawaii}b and Figure \ref{fig: mean profiles Hawaii}c, respectively. For Golden, one can observe that the transition and winter profiles are distinct, while there is some overlap in the summer season between SH and SM before noon profiles shown in Figure \ref{fig: mean profiles Golden}b.  Interestingly, SM, TM and WH show some overlap in the afternoon, further highlighting the ambiguity from the ED-based clustering. It can be noted that all the low irradiance profiles are distinct. 
}

\colorOpt{red}{
The $\beta$-based clustering for Hawaii shows reasonably separated H and M profiles for all the seasons. Notably, they are not as separated as they were for Golden. There is some overlap that can be observed between SM and WH in the afternoon period. While SL is distinct, interestingly, TL and WL have a significant overlap. For the ED-based clusters, High profiles can be considered to be reasonably distinct. The SM is reporting lower irradiance when compared to TM and WM in a good part of the afternoon. As observed earlier with $\beta$ mean profiles, WL and TL are close to each other, even the SL is quite close to them.
}

\colorOpt{red}{Results from DTW distance-based clustering show three distinct summer profiles, but TH and TM, WH, and WM show significant overlap in Figure \ref{fig: mean profiles Golden}c. Interestingly, TL and WL exhibit significant overlap. Additionally, the peak of SL is higher than WH and WM. These observations indicate the ambiguity and confusion resulting from clusters obtained using DTW distance-based clustering. In case of DTW based clusters obtained for Hawaii shown in Figure \ref{fig: mean profiles Hawaii}c, the mean profiles of SH, WH and TH are distinct and well separated. While WM is distinct, SM and TM are quite close to one another. Interestingly, SL appears to align with the SM and TM profiles. The TL and WL are overlapping for the most part but appear to be isolated from others.
}

\colorOpt{red}{Next, the clustering performance of these methods is quantitatively compared using the metrics discussed in the methodology section in Tables \ref{table: clustering performance score Golden} and \ref{table: clustering performance score Hawaii} for Golden and Hawaii, respectively. As anticipated, the performance of $\beta$-based clustering is consistent for both the locations and all the seasons. The performance of $\beta$-based clustering reported Silh scores closer to 1, a high CH score and low DB score.  The performance scores for ED-based clustering are significantly farther from the preferred values, indicating its limitations and poor clustering performance. The DTW distance-based approach has performed worse than ED-based clustering indicating a significantly poorer clustering performance. These measures support the qualitative observations made using the cluster mean profiles.
}

\tableClusteringPerformanceGolden
\tableClusteringPerformanceHawaii
\colorOpt{red}{The performance metrics obtained using direct clustering are shown in `All Season' rows in Table \ref{table: clustering performance score Golden} and Table \ref{table: clustering performance score Hawaii}. While the $\beta$-based clustering showed a small dip from the seasonal clustering, the performance of the other two clustering strategies suffered significantly. Specifically, when direct clustering was applied with ED and DTW distance-based methods, the Silh score came out as -0.07 and 0.03, respectively, in comparison to 0.6 with $\beta$-based clustering for Golden. Similar notable drop was observed for Hawaii. The performance degradation can also be observed in CH and DB scores for both the locations.
}

\colorOpt{red}{Furthermore, noting that almost the same number of days are in a season at both the locations, $\beta$-based clustering identifies a greater number of high $\beta$ days for Hawaii (refer Table \ref{table: clustering performance score Hawaii}) compared to Golden (refer Table \ref{table: clustering performance score Golden}), as expected. ED and DTW-distance-based clustering performed consistently poorly by not picking out as many clear (High) days. This results in the ambiguous mean profiles shown in Figures \ref{fig: mean profiles Hawaii}b-c supported by performance metrics in Table \ref{table: clustering performance score Hawaii}. Interestingly, both the approaches performed very poorly for summer, the season with high variability. ED-based approach predicted more days as SH when compared to other methods. As a result, the peak of SH is lesser than TH in Figure \ref{fig: mean profiles Hawaii}b. Similarly, DTW-based approach had challenges in accurately quantifying the number of days of SH, and SL days. The significant ambiguity is also highlighted in Figure \ref{fig: mean profiles Hawaii}c. We shall discuss the confusion matrices later in this section.
}
\colorOpt{red}{The $\beta$ ranges for each season and irradiance levels are presented in Table \ref{table: beta range Golden} and Table \ref{table: beta range Hawaii} for Golden and Hawaii, respectively for the three clustering approaches.} Being a single feature-based clustering approach, $\beta$-based clustering recommended clear non-overlapping boundaries. Clusters identified by time-series methods exhibit overlapping $\beta$ ranges causing ambiguities. One key observation seems to be that ED based approach located the lower bound of $\beta$ more reliably for each cluster better than the DTW based approach, while both had difficulties in avoiding clear days in Medium or Low irradiance clusters.

\tableBetaRangeGolden

\tableBetaRangeHawaii

\colorOpt{red}{Confusion matrices provide a simple but effective quantitative way to compare the performance of classification methods. The confusion matrices of ED and DTW distance-based approaches are shown in Tables \ref{tab: confusion matrix Golden} and \ref{tab: confusion matrix Hawaii} for Golden and Hawaii, respectively. The entries summarize the probability of correct and incorrect predictions taking $\beta$-based results as reference. The diagonal entries represent the fraction of days that were identified correctly, while the off-diagonal entries represent the mismatched classification. While both the methods have misclassified one class up or down, for example, H as M, M as H, or M as L, DTW distance-based method sometimes classifies some of the L as H or vice versa. Similar results were obtained for both transition and summer seasons. The poor performance of DTW distance-based approach for summer of Hawaii is obvious. Overall, this comparison emphasizes that, for both the locations and for all the seasons, ED-based clustering performed better than DTW distance-based clustering. While DTW method that maps one to many points might be good at identifying similar time-series patterns, overall, the more direct point-to-point mapping in ED method seems to be working better for irradiance clustering. These evaluations provide good insights into the strengths and weaknesses of the clustering approaches in predicting irradiance levels across various seasons.
}

\tableConfusionMatrixGolden

\tableConfusionMatrixHawaii

As highlighted earlier, clustering analysis can form a strong foundation for forecasting models. Next, in this context we explored how strongly a irradiance level of a day is influenced by the irradiance leel of the previous day captured by the probabilistic distributions of the following day given the previous day's irradiance level. The variation of these distributions with seasons for Golden using $\beta$-based clusters is discussed here.

\figureSeasonalSecondDayTransitionGoldenHawaii


\colorOpt{red}{
The transition probabilities for the three seasons are shown in Figures \ref{fig: seasonal consecutive days transition Golden Hawaii}a and \ref{fig: seasonal consecutive days transition Golden Hawaii}b for Golden and Hawaii, respectively. The figures reveal intriguing patterns in the daily weather transitions. For example, in Golden, during summer, the probability of a clear day (H) following another clear day is only 50\%, on the other hand for the transition season, it increases to 68\%. Notably, the likelihood of transitioning from a clear day (H) to a cloudy day (L) remains low across all seasons, but the reverse of transitioning from a cloudy day to a clear day shows higher probability especially in the winter. In the winter, the probability of transitioning to a clear day is the highest among all day types, indicating that weather variability is generally lower in winter compared to other seasons, with weather transitions tending to be more gradual. 
}

\colorOpt{red}{
Analysis of transition probabilities shown in Figure \ref{fig: seasonal consecutive days transition Golden Hawaii}b for Hawaii reveals even higher probabilities of a clear day following another clear day, or even a partially cloudy (M) or cloudy day, compared to Golden. For example, in winter, the probability of transitioning from a clear day to another clear day is 82\%, which is significantly higher than the 57\% observed for Golden. Similarly, in summer, this probability is 64\% in Hawaii compared to 50\% for Golden. However, an exception is noted for winter, where the probability of transitioning from a cloudy day to a clear day is 60\% in Golden but only 29\% for Hawaii. Interestingly, the probability of a clear day to a cloudy day transition in winter has significantly dropped from 15\% to 3\%. This suggests a strong persistence of clear weather patterns in Hawaii during the winter, reflecting the region's relatively stable and predictable weather conditions in this season. 
}

\colorOpt{red}{
These observations underscore the complex interplay between consecutive days' irradiance patterns, influenced by seasonal and geographical factors. These findings not only highlight the distinctive weather dynamics of Golden and Hawaii but also provide valuable insights that can guide the development of tailored forecasting models for solar power or irradiance forecasting. By understanding these transition probabilities, it would be beneficial to develop season-specific forecasting models customized to the specific weather characteristics of different locations, contributing to a more reliable, efficient and better planning and optimization of solar energy systems.
}

\section{Summary and Conclusion}
\label{conclusion}

Solar irradiance clustering serves as a pivotal tool, offering a quantitative assessment of solar power generation potential for a site. By identifying typical irradiance patterns, it contributes to the development of accurate forecasting models crucial for enhanced grid integration and growth of global solar power installed capacity.

\colorOpt{red}{We have adopted a two-level clustering approach to comprehensively analyze solar irradiance leading to a good understanding of the typical weather/cloud conditions pertinent to two representative sites. In the first level, well-defined contiguous solar seasons were identified using clear sky irradiance for the site with a year-on-year repeating patterns of winter, transition, summer, and transition. In the second level, to capture the irradiance level in a day, the daily irradiance index (DII or $\beta$) recommended by Agrawal et al. \citep{rosh2023} was used. The efficacy of this index was first established by comparing it with the measured total cloud cover using the data published by NREL for Golden, Colorado \citep{nrel}.}

In the second level of clustering, feature based clustering using $\beta$, and time series distance-based clustering using ED and DTW were performed to identify three irradiance levels in a season as High, Medium and Low. The quality of clustering methods was quantified using three scores, namely the Silhouette score, Calinski-Harabasz score, and Davies-Bouldin score.  The $\beta$-based approach found to be superior to the popular time-series clustering techniques based on these scores. This study supports the robustness, efficiency, and wide applicability of the $\beta$-based approach for irradiance clustering. Taking $\beta$ based predictions as the ground truth, confusion matrices were obtained for the time-series clustering techniques. Interestingly, among the time series distance-based methods, contrary to the expectations ED performed better. 

\colorOpt{red}{The transition probabilities of seasons based on irradiance levels could be used to make high-level forecasting of solar irradiance that could be useful for generating realistic input data for PV simulators with different irradiance patterns typical for the location. Future work could focus on building robust forecasting models for $\beta$-based clusters for various seasons.}

\section*{Acknowledgement}
The authors would like to thank the NREL for sharing the data with open access. The authors acknowledge and appreciate the support of Dr. Harrick Vinn, CTO, Tata Consultancy Services
(TCS), Dr. Gautam Shroff, former Head, TCS Research, and Dr. Sachin Lodha, Head, TCS Research in pursuing this research.



\bibliographystyle{elsarticle-num} 
\bibliography{bibliography}

\begin{thebibliography}{10}
\expandafter\ifx\csname url\endcsname\relax
  \def\url#1{\texttt{#1}}\fi
\expandafter\ifx\csname urlprefix\endcsname\relax\def\urlprefix{URL }\fi
\expandafter\ifx\csname href\endcsname\relax
  \def\href#1#2{#2} \def\path#1{#1}\fi

\bibitem{Ourworld}
H.~Ritchie, M.~Roser, P.~Rosado, Renewable energy, Our World in {Data.
  }{https://ourworldindata.org/renewable-energy} (2023).

\bibitem{kim2023}
E.~Kim, M.~S. Akhtar, O.-B. Yang, Designing solar power generation output
  forecasting methods using time series algorithms, Electric Power Systems
  Research 216 (2023) 109073.

\bibitem{paletta2021}
Q.~Paletta, G.~Arbod, J.~Lasenby, Benchmarking of deep learning irradiance
  forecasting models from sky images--an in-depth analysis, Solar Energy 224
  (2021) 855--867.

\bibitem{chun2017b}
C.~S. Lai, Y.~Jia, M.~D. McCulloch, Z.~Xu, Daily clearness index profiles
  cluster analysis for photovoltaic system, IEEE Transactions on Industrial
  Informatics 13~(5) (2017) 2322--2332.

\bibitem{baor2018}
B.~Zhou, W.~Zhao, X.~Su, S.~Lu, T.~Wang, W.~Yao, P.~Xie, T.~Mao, L.~Guan,
  Y.~Lv, {PV} power characteristic modeling based on multi-scale clustering and
  its application in generation prediction, in: 2018 IEEE power \& energy
  society general meeting (PESGM), IEEE, 2018, pp. 1--5.

\bibitem{paul2018}
P.~Govender, M.~J. Brooks, A.~P. Matthews, Cluster analysis for classification
  and forecasting of solar irradiance in durban, south africa, Journal of
  Energy in Southern Africa 29~(2) (2018) 51--62.

\bibitem{hui2018}
H.~He, R.~Hu, Y.~Zhang, Y.~Zhang, R.~Jiao, A power forecasting approach for
  {PV} plant based on irradiance index and lstm, in: 2018 37th Chinese control
  conference (CCC), IEEE, 2018, pp. 9404--9409.

\bibitem{spyr2020}
S.~Theocharides, G.~Makrides, A.~Livera, M.~Theristis, P.~Kaimakis, G.~E.
  Georghiou, Day-ahead photovoltaic power production forecasting methodology
  based on machine learning and statistical post-processing, Applied Energy 268
  (2020) 115023.

\bibitem{maoYang2022}
M.~Yang, M.~Zhao, D.~Huang, X.~Su, A composite framework for photovoltaic
  day-ahead power prediction based on dual clustering of dynamic time warping
  distance and deep autoencoder, Renewable Energy 194 (2022) 659--673.

\bibitem{paul2019}
P.~Govender, V.~Sivakumar, Investigating diffuse irradiance variation under
  different cloud conditions in durban, using k-means clustering, Journal of
  Energy in Southern Africa 30~(3) (2019) 22--32.

\bibitem{isab2021}
I.~Santiago, J.~L. Esquivel-Martin, D.~Trillo-Montero, R.~J. Real-Calvo,
  V.~Pallar{\'e}s-L{\'o}pez, Classification of daily irradiance profiles and
  the behaviour of photovoltaic plant elements: The effects of cloud
  enhancement, Applied Sciences 11~(11) (2021) 5230.

\bibitem{fahd2020}
F.~Amjad, L.~A. Shah, Identification and assessment of sites for solar farms
  development using gis and density based clustering technique-a case of
  pakistan, Renewable Energy 155 (2020) 761--769.

\bibitem{josh2013}
J.~Traube, F.~Lu, D.~Maksimovic, J.~Mossoba, M.~Kromer, P.~Faill, S.~Katz,
  B.~Borowy, S.~Nichols, L.~Casey, Mitigation of solar irradiance intermittency
  in photovoltaic power systems with integrated electric-vehicle charging
  functionality, IEEE Transactions on Power Electronics 28~(6) (2013)
  3058--3067.

\bibitem{lauret2022}
P.~Lauret, R.~Alonso-Su{\'a}rez, J.~Le~Gal La~Salle, M.~David, Solar forecasts
  based on the clear sky index or the clearness index: Which is better?, in:
  Solar, Vol.~2, MDPI, 2022, pp. 432--444.

\bibitem{zhang2021}
Z.~Zhang, C.~Wang, X.~Peng, H.~Qin, H.~Lv, J.~Fu, H.~Wang, Solar radiation
  intensity probabilistic forecasting based on k-means time series clustering
  and gaussian process regression, IEEE Access 9 (2021) 89079--89092.

\bibitem{cong2019a}
C.~Feng, M.~Cui, B.-M. Hodge, S.~Lu, H.~F. Hamann, J.~Zhang, An unsupervised
  clustering-based short-term solar forecasting methodology using multi-model
  machine learning blending, arXiv preprint arXiv:1805.04193 (2018).

\bibitem{chun2017a}
C.~S. Lai, X.~Li, L.~L. Lai, M.~D. McCulloch, Daily clearness index profiles
  and weather conditions studies for photovoltaic systems, Energy Procedia 142
  (2017) 77--82.

\bibitem{ebuka2021}
E.~B. Osunwoke, S.~S. Ullah, A.~J. Abianeh, F.~Ferdowsi, T.~L. Chambers, A
  machine learning-enabled clustering approach for large-scale classification
  of solar data, in: 2021 North American Power Symposium (NAPS), 2021, pp.
  01--06.

\bibitem{tanaka2016}
Y.~Tanaka, M.~Takahashi, Dynamic time warping-based cluster analysis and
  support vector machine-based prediction of solar irradiance at multi-points
  in a wide area, in: Proceedings of the ISCIE International Symposium on
  Stochastic Systems Theory and its Applications, Vol. 2016, The ISCIE
  Symposium on Stochastic Systems Theory and Its Applications, 2016, pp.
  210--215.

\bibitem{take2016}
T.~Watanabe, T.~Takamatsu, T.~Y. Nakajima, Evaluation of variation in surface
  solar irradiance and clustering of observation stations in japan, Journal of
  Applied Meteorology and Climatology 55~(10) (2016) 2165--2180.

\bibitem{maldonado2022}
P.~Maldonado-Salguero, M.~C. Bueso-S{\'a}nchez, {\'A}.~Molina-Garc{\'\i}a,
  J.~M. S{\'a}nchez-Lozano, Spatio-temporal dynamic clustering modeling for
  solar irradiance resource assessment, Renewable Energy 200 (2022) 344--359.

\bibitem{xing2018}
X.~Fu, F.~Gao, J.~Wu, X.~Guan, X.~Li, P.~Liu, P.~Li, A simulation method of
  solar irradiance data based on feature clustering and markov transition
  probability matrix, in: 2018 13th World Congress on Intelligent Control and
  Automation (WCICA), IEEE, 2018, pp. 1741--1746.

\bibitem{shua2017}
S.~Li, H.~Ma, W.~Li, Typical solar radiation year construction using k-means
  clustering and discrete-time markov chain, Applied Energy 205 (2017)
  720--731.

\bibitem{patr2013}
P.~Jeanty, M.~Delsaut, L.~Trovalet, H.~Ralambondrainy, J.~Lan-Sun-Luk,
  M.~Bessafi, P.~Charton, J.-P. Chabriat, Clustering daily solar radiation from
  reunion island using data analysis methods, in: Proceedings of the
  International Conference on Renewable Energies and Power Quality (ICREPQ'13),
  Bilbao, Spain, Vol. 340, 2013.

\bibitem{azed2019}
{\^A}.~Frimane, W.~A. Ahmed, B.~Ouhammou, M.~Aggour, Automatic classification
  of daily-histograms of solar irradiance in morocco: A nonparametric-bayesian
  paradigm, in: Proceedings of the New Challenges in Data Sciences: Acts of the
  Second Conference of the Moroccan Classification Society, 2019, pp. 1--5.

\bibitem{rosh2023}
R.~Agrawal, S.~Subramanian, S.~Agarwal, V.~Runkana, A robust approach for daily
  solar irradiance clustering, in: 2023 IEEE 50th Photovoltaic Specialists
  Conference (PVSC), 2023, pp. 1--6.

\bibitem{ribeiro2024}
R.~Ribeiro, B.~Fanzeres, Identifying representative days of solar irradiance
  and wind speed in brazil using machine learning techniques, Energy and AI 15
  (2024) 100320.

\bibitem{take2021}
T.~Watanabe, K.~Oka, Y.~Hijioka, Assessment of characteristics of surface solar
  irradiance on consecutive days using a self-organizing map and clustering
  methods, Meteorological Applications 28~(2)  e1984.

\bibitem{javier2022}
J.~Lopez-Lorente, S.~Theocharides, G.~Makrides, G.~E. Georghiou, Impact of
  daily irradiance profiles on intra-day solar forecasting, in: 2022 IEEE 49th
  Photovoltaics Specialists Conference ({PVSC}), 2022, pp. 0156--0163.

\bibitem{MuMo2016}
A.~A. Munshi, Y.~A.-R. Mohamed, Photovoltaic power pattern clustering based on
  conventional and swarm clustering methods, Solar Energy 124 (2016) 39--56.

\bibitem{chris2013}
C.~Trueblood, S.~Coley, T.~Key, L.~Rogers, A.~Ellis, C.~Hansen, E.~Philpot,
  {PV} measures up for fleet duty : Data from a tennessee plant are used to
  illustrate metrics that characterize plant performance, IEEE Power and Energy
  Magazine 11~(2) (2013) 33--44.

\bibitem{bali2020}
B.~Hartmann, Comparing various solar irradiance categorization methods--a
  critique on robustness, Renewable Energy 154 (2020) 661--671.

\bibitem{scikit-learn}
F.~Pedregosa, G.~Varoquaux, A.~Gramfort, V.~Michel, B.~Thirion, O.~Grisel,
  M.~Blondel, P.~Prettenhofer, R.~Weiss, V.~Dubourg, J.~Vanderplas, A.~Passos,
  D.~Cournapeau, M.~Brucher, M.~Perrot, E.~Duchesnay, Scikit-learn: Machine
  learning in {P}ython, Journal of Machine Learning Research 12 (2011)
  2825--2830.

\bibitem{de2018}
D.~G. de~Barros~Franco, M.~T.~A. Steiner, Clustering of solar energy facilities
  using a hybrid fuzzy c-means algorithm initialized by metaheuristics, Journal
  of cleaner production 191 (2018) 445--457.

\bibitem{nrel}
T.~Stoffel, A.~Andreas, {NREL} solar radiation research laboratory {SRRL}:
  Baseline measurement system {BMS}; golden, colorado (data), {NREL} Data
  Catalog. Golden, Colorado (Data); {NREL} Report No.
  DA-5500-56488.Https://midcdmz.nrel.gov/apps/day.pl?BMS.

\bibitem{nrelnelha}
A.~Olson, K.;~Andreas, Natural energy laboratory of hawaii authority ({NELHA}):
  Hawaii ocean science \& technology park; kailua-kona, hawaii (data), NREL
  Report No. DA-5500-64450Http://dx.doi.org/10.7799/1183461.

\bibitem{ineichen2008}
P.~Ineichen, A broadband simplified version of the solis clear sky model, Solar
  Energy 82~(8) (2008) 758--762.

\bibitem{pvlib}
W.~F. Holmgren, C.~W. Hansen, M.~A. Mikofski, {PVLIB} python: A python package
  for modeling solar energy systems, Journal of Open Source Software 3~(29)
  (2018) 884.

\end{thebibliography}




\end{document}